\documentclass[twoside]{article}

\usepackage[accepted]{aistats2022}
\usepackage[round]{natbib}

\usepackage{siunitx}
\usepackage{hyperref} 
\usepackage{xcolor}
\hypersetup{
    colorlinks,
    linkcolor={red!50!black},
    citecolor={blue!50!black},
    urlcolor={blue!80!black}
}
\usepackage{algorithm}
\usepackage{algorithmic}
\usepackage{paralist}
\usepackage{microtype}
\usepackage{graphicx}
\usepackage{subcaption}
\usepackage{booktabs} 
\usepackage{amsmath}
\usepackage{amsthm}
\usepackage{mathtools}
\usepackage{bm}
\usepackage{amssymb}
\DeclareMathOperator*{\argmax}{arg\,max}
\usepackage{dsfont}
\newtheorem{proposition}{Proposition}
\newtheorem{definition}{Definition}
\newtheorem{lemma}{Lemma}

\newtheorem*{remark}{Remark}
\usepackage{multirow}
\usepackage{tikz}
\usepackage{environ}
\usepackage{adjustbox}
\usepackage{pgfmath, pgffor}
\usepackage{cancel}
\usetikzlibrary{math,calc}
\usetikzlibrary{positioning}
\usetikzlibrary{decorations.pathreplacing, arrows.meta,spy,shapes.geometric}
\usetikzlibrary{fit}
\usepackage{standalone}
\usepackage{dcolumn}
\DeclareMathVersion{nxbold}
\SetSymbolFont{operators}{nxbold}{OT1}{cmr} {b}{n}
\SetSymbolFont{letters}  {nxbold}{OML}{cmm} {b}{it}
\SetSymbolFont{symbols}  {nxbold}{OMS}{cmsy}{b}{n}
\usepackage{collcell}
\newcolumntype{.}{D{.}{.}{-1}}
\newcolumntype{d}[1]{D{.}{.}{#1}}
\newcommand{\boldunderlined}[1]{\mathversion{nxbold}\underline{#1}}
\makeatletter
\newcolumntype{U}[3]{>{\collectcell\boldunderlined\DC@{#1}{#2}{#3}}c<{\DC@end\endcollectcell}}
\newcolumntype{B}[3]{>{\boldmath\DC@{#1}{#2}{#3}}c<{\DC@end}}
\newcolumntype{Z}[3]{>{\mathversion{nxbold}\DC@{#1}{#2}{#3}}c<{\DC@end}}
\makeatother

\usepackage{mathtools}

\DeclarePairedDelimiter\abs{\lvert}{\rvert}%
\DeclarePairedDelimiter\norm{\lVert}{\rVert}%

\makeatletter
\let\oldabs\abs
\def\abs{\@ifstar{\oldabs}{\oldabs*}}
\let\oldnorm\norm
\def\norm{\@ifstar{\oldnorm}{\oldnorm*}}
\makeatother
\begin{document}

%

%

\twocolumn[

\aistatstitle{Deep Neyman-Scott Processes}

\aistatsauthor{ Chengkuan Hong \And Christian R. Shelton }

\aistatsaddress{ UC Riverside\\chong009@ucr.edu \And UC Riverside\\cshelton@cs.ucr.edu } ]

\begin{abstract}
  A Neyman-Scott process is a special case of a Cox process. The latent and
observable stochastic processes are both Poisson processes. We consider a
deep Neyman-Scott process in this paper, for which the building components
of a network are all Poisson processes. We develop an efficient posterior
sampling via Markov chain Monte Carlo and use it for likelihood-based
inference. Our method opens up room for the inference in sophisticated
hierarchical point processes. We show in the experiments that more hidden
Poisson processes brings better performance for likelihood fitting and
events types prediction.  We also compare our method with state-of-the-art
models for temporal real-world datasets and demonstrate competitive
abilities for both data fitting and prediction, using far fewer parameters.
\end{abstract}

\section{INTRODUCTION}

Point processes have attracted attention due to their ability to model the
temporal and spatial patterns of event data. They have been applied to
various fields, \textit{e.g.}, finance \citep{bauwens2009modelling}, neuroscience
\citep{perkel1967neuronal}, and cosmology \citep{stoica2014}.  The Poisson 
process is one of the most commonly used models. The intensity of a
Poisson process describes the expected number of events per unit time or
space and is independent of the events elsewhere.

We usually do not have any prior knowledge of the functional form of the
intensity function. A common strategy is to assume the intensity function
itself is a latent stochastic process.  By introducing this latent process, we
introduce dependence between the expected number of events at different times or
places.
With this strategy, we have a class
of point processes called Cox processes \citep{cox1955some}. Typical
examples for the latent stochastic process are Poisson processes
\citep{neyman1958statistical}, Cox processes
\citep{adams2009kernel}, Strauss processes 
\citep{yau2012generalization}, and Gaussian processes 
\citep{adams2009tractable}.

In this paper, we consider a special class of Cox processes, Neyman-Scott
processes (N-SPs) of order
larger than one \citep{neyman1958statistical}, which we call deep
Neyman-Scott processes (DN-SPs). Neyman and Scott first described univariate DN-SPs in
1958; they were trying to model the distribution of galaxies in the
Universe.  Each building block of the univariate DN-SPs is a Poisson process. Taking
univariate DN-SPs of order two as an example, one Poisson process generates the
centers of some superclusters, consisting of galaxy clusters.
In turn, each galaxy cluster generates its own set of points, galaxies,
from a Poisson process centered the cluster center. The advantage of the DN-SPs compared to the other Cox processes is that they have the ability to build
deep hierarchical models where each component in the system is a point
process.  Further, unlike Hawkes processes \citep{Haw71} and similar temporal processes,
DN-SP processes do not enforce a temporal or causal ``direction,'' and thus also encompass
spatial and spatio-temporal processes.

\begin{figure*}[htb]
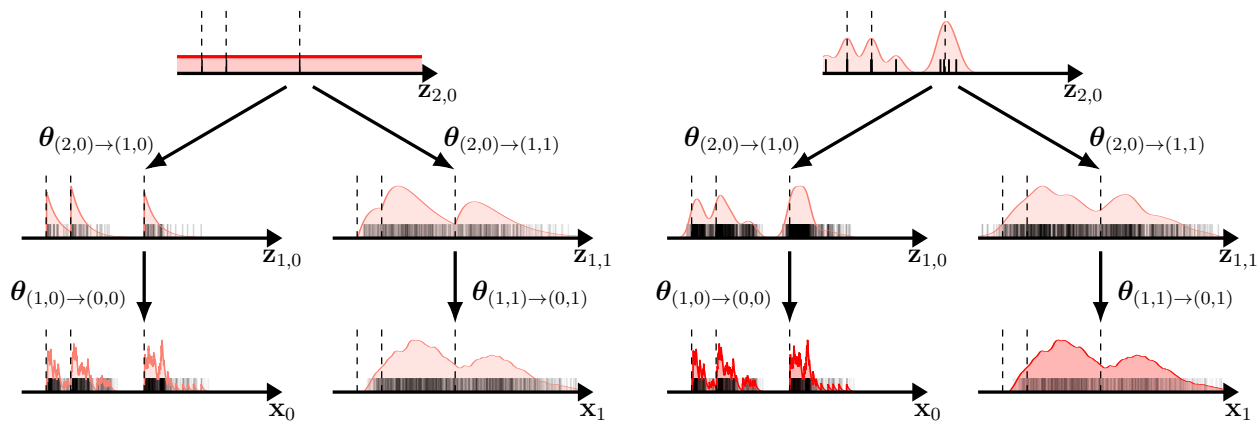

\centering
\begin{subfigure}[b]{0.48\textwidth}
\includestandalone[width=\textwidth]{dnsp_forward_sampling}
\end{subfigure}
~
\begin{subfigure}[b]{0.48\textwidth}
\includestandalone[width=\textwidth]{dnsp_posterior_sampling}
\end{subfigure}
\vspace{.15in}
	\caption{Illustration of Model Sampling: (left) forward sampling \& (right) posterior sampling.}
\label{fig:dpp-ex} \vskip -0.2in
\end{figure*}

N-SPs of order of one have been well studied in the literature.
\citet{moller2003statistical} discuss posterior sampling algorithms 
and inference methods for univariate N-SPs based on the Metropolis-Hastings (M-H)
\citep{metropolis1953equation,norman1969investigations,hastings1970monte}
and spatial birth-and-death (SB\&D)
\citep{kelly1976note,ripley1977modelling,baddeley1989nearest}.
\citet{linderman2017bayesian} consider a sequential Monte Carlo sampler for
a multivariate (not deep) N-SP, but without giving a functional form for the proposal.
\citet{williams2020point} give a collapsed Gibbs posterior sampling
algorithm for a univariate (not deep) N-SP. However, little attention has
been given to DN-SPs (order greater than one) due to the difficulty in the complex posterior
sampling and inference. The only existing work we have found is from \citet{andersen2018double}, which
employs a moment-based inference for photoactivated localization microscopy. Although the moment-based
inference is  computationally easy, some constant parameters for the estimation of the moments have to
be tuned manually, and it is not able to estimate the posterior intensity surface of the unobservable point processes.

To the best of our knowledge, we are the first to
give a posterior sampling algorithm and a likelihood-based inference method for multivariate DN-SPs with experiments for real-world data. In this paper, we first
construct a temporal multivariate DN-SP in one dimension in Sec.~\ref{sec:DN-SP-model}.  Then in Sec.~\ref{sec:inference}, we present an efficient posterior sampling via Markov chain Monte Carlo (MCMC) and use it
for inference. We give competitive experiments results for real-world data
compared with MTPP \citep{lian2015multitask}, the neural Hawkes process \citep{mei2017neural}, the
self-attentive Hawkes process \citep{zhang2020self}, and the
transformer Hawkes process \citep{zuo2020transformer} in
Sec.~\ref{sec:experiments}. While our experiments are for temporal point processes, our inference method can also be applied to general spatial point processes as discussed in Sec.~\ref{sec:mcmc}.

\section{DEEP NEYMAN-SCOTT PROCESSES}
\label{sec:DN-SP-model}

In this section, we review temporal point processes and introduce our DN-SPs. Different from the N-SPs
given by \citet{neyman1958statistical}, our DN-SPs are multivariate (events can be marked or labeled).

\paragraph{Temporal Point Processes}
A temporal point process (TPP) \citep{theofppvo1} is a random process with a
realization as a sequence of events \(\{t_i\}_{i=1}^{m}\), where each time
point \(t_i \in \mathbb{R}_{\geq 0}\) and \(t_i < t_{i+1}, \forall i\).
\begin{definition}[conditional intensity function] The {\em conditional
	intensity function} (CIF) of a TPP is defined as 
\begin{align*}
	\lambda(t)=\lim_{\Delta t\rightarrow 0}\frac{\text{\em Pr}(\text{\em One event in }[t,t+\Delta t]\ \mid\ \mathcal{H}_t)}{\Delta t},
\end{align*}
where \(\mathcal{H}_t\) is an element in a filtration \(\{\mathcal{H}_t\}_{t \geq 0}\).
\end{definition}
TPPs are completely characterized by their CIFs as they determine the
distributions of the numbers of events as Bernoulli random variables at
each infinitesimal interval. As a special case for TPPs, a homogeneous
Poisson process (HPP) has a constant CIF \(\lambda_0 > 0\). At each
infinitesimal interval \([t,t+\Delta t]\), the probability of one
event is \(\lambda_0 \cdot \Delta t\).

\paragraph{Deep Neyman-Scott Processes} To build DN-SPs, we will stack TPPs
in a hierarchical manner \textit{s.t.} the distributions of the CIFs as
random processes are controlled by the TPPs on higher layers.

Each observed data point \(\mathbf{x}\) is a collection of sequences
\(\{\mathbf{x}_{k}\}_{k=1}^{K_0}=\{\{t_{0,k,j}\}_{j=1}^{m_{0,k}}\}_{k=1}^{K_0}\),
where \(\mathbf{x}_k\) is the sequence of the \(k\)th type of event (or events with the
\(k\)th mark in a discrete-marked TPP) and \(t_{0,k,j}\) is the time of the
\(j\)th 
event of this type.  (The \(0\) indicates this is an observation event at the
bottom layer of the model.)
For each data point \(\mathbf{x}\), there are \(L\) hidden
layers of TPPs \(\mathbf{Z}=\{\mathbf{Z}_{1}, \dots, \mathbf{Z}_{L}\}\),
with \(\mathbf{Z}_{\ell}=\{Z_{\ell,k}\}_{k=1}^{K_\ell}\).  \(K_\ell\) is
the number of hidden processes at level \(\ell\), and  \(Z_{\ell,k}\)
is a TPP. \(Z_{\ell,k}\) and \(Z_{\ell+1,j}\) are connected by a kernel
function \(\phi_{\bm{\theta}_{(\ell+1,k) \rightarrow (\ell,j)}}(\cdot)\),
whose functional form will be given later.

\paragraph{Generative Model Semantics}
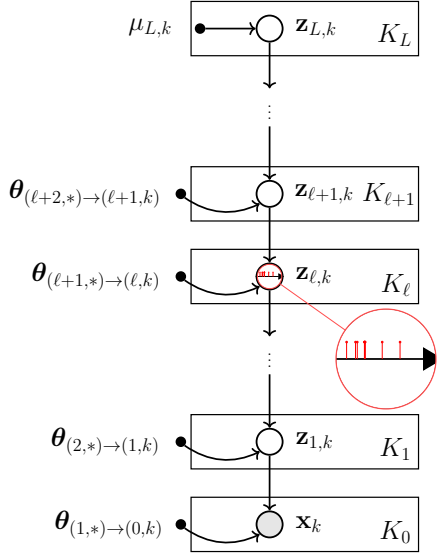
\begin{figure}[htb]
\vspace{.2in}
\begin{center}
\centering
	\scalebox{1.0}{\begin{tikzpicture}
[
recnode/.style={      rectangle,
      draw=black,
      thick,
      fill=white,
      text width=5cm,
      align=right,
      minimum height=1.25cm},
recnode2/.style={      rectangle,
      thick,
      text width=5cm,
      align=right,
      minimum height=1.25cm},   
recnode3/.style={      rectangle,
      draw=black,
      thick,
      text width=5cm,
      minimum height=1.25cm},  
circlenode/.style={circle, draw=black, very thick, minimum size=6mm},
circlenode2/.style={circle, very thick, minimum size=6mm},
circlenodewithevents/.style={circle,draw=black, very thick, minimum size=2.5cm},
textnode1/.style={circle, very thick, minimum size=6mm},
textnode2/.style={rectangle},
blacknode/.style={circle,draw=black, fill=black, very thick, minimum size=2mm, inner sep=0pt},
spy using outlines={circle, magnification=4.1, connect spies}
]
\node[recnode] (1) {};
\node[recnode2,below=0.625cm of 1.south east, anchor=north east] (2) {};
\node[recnode3,below=0.625cm of 2.south east, anchor=north east] (3) {};
\node[recnode3,below=0.625cm of 3.south east, anchor=north east] (4) {};
\node[recnode2,below=0.625cm of 4.south east, anchor=north east] (5) {};
\node[recnode3,below=0.625cm of 5.south east, anchor=north east] (6) {};
\node[recnode3,below=0.625cm of 6.south east, anchor=north east] (7) {};
\node[circlenode, right=1.5cm of 3.west] (8) {};
\node[textnode1,right=0.05cm of 8.east] (15) {\LARGE \(\mathbf{z}_{\ell+1,k}\)};
\node[blacknode, left=1mm of 3.west] (22) {};
\node[textnode1,left=2cm of 8.west] (29) {\LARGE \(\bm{\theta}_{(\ell+2,*)\rightarrow(\ell+1,k)}\)};

\node[circlenode2,above=1.875cm of 8.south] (9) {\(\vdots\)};

\node[circlenode,right=1.5cm of 1.west] (10) {};
\node[textnode1,right=0.05cm of 10.east] (16) {\LARGE \(\mathbf{z}_{L,k}\)};
\node[blacknode, right=1mm of 1.west] (21) {};
\node[textnode1,left=3mm of 21.west] (20) {\LARGE \(\mu_{L,k}\)};

\node[circlenode, right=1.5cm of 4.west] (11) {};
\node[textnode1,right=0.05cm of 11.east] (17) {\LARGE \(\mathbf{z}_{\ell,k}\)};
\node[blacknode, left=1mm of 4.west] (23) {};
\node[textnode1,left=2cm of 11.west] (26) {\LARGE \(\bm{\theta}_{(\ell+1,*)\rightarrow(\ell,k)}\)};

\draw[-Triangle,line width=0.1mm] (11.west) -- (11.east);
\foreach \i in {1mm,1.5mm,1.6mm,2mm,2.01mm,2.02mm,2.03mm,2.04mm,3mm,4mm}
\draw[red,-{Circle[length=0.1mm,width=0.1mm]},line width=0.01mm] ($(11.west)!\i!(11.east)$) -- ($(11.west)!\i!(11.east)+(0,0.1)$);
\spy [red!70,size=2.3cm] on (11) in node at ($(17.east)+(1,-1.9)$);

\node[circlenode, right=1.5cm of 6.west] (13) {};
\node[circlenode2, above=1.875cm of 13.south] (12) {\(\vdots\)};
\node[textnode1,right=0.05cm of 13.east] (18) {\LARGE \(\mathbf{z}_{1,k}\)};
\node[blacknode, left=1mm of 6.west] (24) {};
\node[textnode1,left=2cm of 13.west] (28) {\LARGE \(\bm{\theta}_{(2,*)\rightarrow(1,k)}\)};

\node[circlenode,fill=black!10, right=1.5cm of 7.west] (14) {};
\node[textnode1,right=0.05cm of 14.east] (19) {\LARGE \(\mathbf{x}_{k}\)};
\node[blacknode, left=1mm of 7.west] (25) {};
\node[textnode2,left=2cm of 14.west] (27) {\LARGE \(\bm{\theta}_{(1,*)\rightarrow(0,k)}\)};

\draw[->,line width=0.4mm] (21.east) -- (10.west);
\draw[->,line width=0.4mm] (10.south) -- (9.north);
\draw[->,line width=0.4mm] (9.south) -- (8.north);
\draw[->,line width=0.4mm] (8.south) -- (11.north);
\draw[->,line width=0.4mm] (11.south) -- (12.north);
\draw[->,line width=0.4mm] (12.south) -- (13.north);
\draw[->,line width=0.4mm] (13.south) -- (14.north);

\draw[->,line width=0.4mm,bend right=90] (22.south east) to [out=-30,in=-150] (8.south west);
\draw[->,line width=0.4mm,bend right=90] (23.south east) to [out=-30,in=-150] (11.south west);
\draw[->,line width=0.4mm,bend right=90] (24.south east) to [out=-30,in=-150] (13.south west);
\draw[->,line width=0.4mm,bend right=90] (25.south east) to [out=-30,in=-150] (14.south west);

\node[textnode1, above left=0.2mm and 1 mm of 1.south east] {\LARGE \(K_L\)};
\node[textnode1, above left=0.2mm and 1 mm of 3.south east] {\LARGE \(K_{\ell+1}\)};
\node[textnode1, above left=0.2mm and 1 mm of 4.south east] {\LARGE \(K_{\ell}\)};
\node[textnode1, above left=0.2mm and 1 mm of 6.south east] {\LARGE \(K_{1}\)};
\node[textnode1, above left=0.2mm and 1 mm of 7.south east] {\LARGE \(K_0\)};

\end{tikzpicture}}
\vspace{.15in}
\caption{The Structure of A DN-SP} 
\label{fig:dppstructure}
\end{center}
\vskip -0.2in
\end{figure}
We first draw samples from the TPPs on the top layer. These TPPs are assumed to be HPPs. The CIF for \(Z_{L,k}\) is a constant function 
\begin{equation*}
    \lambda_{L,k}(t)=\mu_{k}, \text{ where }\mu_{k} > 0.
\end{equation*}
Next, we draw samples for each hidden TPPs conditional on the TPPs in the layer immediately above. The CIF for \(Z_{\ell,k}\) conditional on \(\mathbf{Z}_{\ell+1}\) is
\begin{equation}
    \lambda_{\ell,k}(t)=\sum_{i=1}^{K_{\ell+1}}\sum_{t_{\ell+1,i,j}}\phi_{\bm{\theta}_{(\ell+1,i) \rightarrow (\ell,k)}}(t-t_{\ell+1,i,j}), \label{eq:intensity-hidden}
\end{equation}
where
\(\mathbf{z}_{\ell+1,i} =
\{t_{\ell+1,i,j}\}_{j=1}^{m_{\ell+1,i}}\)
is a realization for  \(Z_{\ell+1,i}\).

As an example,
Fig.~\ref{fig:dpp-ex} demonstrates the distributions of the events for a DN-SP with
two hidden layers. The left figure is for forward sampling and the right one is for
posterior sampling. For forward sampling, three events are
	drawn from \(Z_{2,0}\).
	The dashed lines indicate the positions of
	the three events on the top layer.
	The plots for the other TPPs are the
densities and rug plots of the samples drawn conditioned on \(\mathbf{z}_{2,0}\).
	For posterior sampling, the samples for \(\mathbf{x}_0\) and \(\mathbf{x}_1\)
are collected and fixed from the forward sampling. Then we draw posterior
samples for the other TPPs conditional on \(\mathbf{x}_0\) and
\(\mathbf{x}_1\) and plot their densities and rug plots as well. Note
the modes of the posterior distribution for \(\mathbf{z}_{2,0}\)
recover the positions of the prior events for \(\mathbf{z}_{2,0}\) in
forward sampling.

Fig.~\ref{fig:dppstructure} gives the general
structure of a DN-SP similar to a graphical model, where each node represents
a TPP,
\(\bm{\theta}_{(\ell+1,*)\rightarrow(\ell,k)}=[\bm{\theta}_{(\ell+1,i)\rightarrow(\ell,k)}]_{i=1}^{K_{\ell+1}}\),
and a sample node for \(\mathbf{z}_{\ell,k}\) is enlarged.

Conditioned on \(\mathbf{Z}_{\ell+1}\), \(Z_{\ell,k}\) is a Poisson process whose CIF at each time
interval is independent of the CIF at the other time intervals, and the number of events at each time interval follows the Poisson distribution. 

The time interval for each hidden TPP is set to be the same as the evidence and no edge effects are considered in our model.

\paragraph{Evidence Likelihood} The observed data are assumed to be drawn
conditioned on the hidden TPPs on the lowest hidden layer in a
manner analogous to those of the layers above.  That is, the intensity
for \(\mathbf{x}_{k}\), \(\lambda_{0,k}(t)\), is as in Eq.~\ref{eq:intensity-hidden}
where \(\ell=0\).
%

\paragraph{The Kernel Function} For a TPP, it is reasonable to make the
kernel function be zero for negative inputs (\textit{i.e.}, to be
``causal'') and converge to \(0\) as \(t\) goes to infinity (%
the influence of the events
will eventually disappear). The sole constrain is
that the sum appearing in the intensity function
(Eq.~\ref{eq:intensity-hidden})
should always be
non-negative. We also want the kernel function to be as flexible as possible with only a few parameters, so we choose a gamma kernel such
that
\begin{equation*}
    \phi_{\bm{\theta}}(x) = \begin{cases}
    p \cdot \frac{\beta^\alpha}{\Gamma(\alpha)}x^{\alpha-1}e^{-\beta x}\text{, for }x>0,\ p,\alpha,\beta > 0,\\
    0,\text{ for }x \leq 0,
    \end{cases}
\end{equation*}
where \(\bm{\theta}=\{p,\alpha,\beta\}\) and \(\Gamma(\alpha)\) is the
gamma function. The gamma kernel asymptotes to \(0\) as \(x\rightarrow \infty\).
Moreover, with the varying parameters, the gamma kernel can either be
monotonically decreasing or have a unimodal shape, which provides flexibility.

\section{INFERENCE} \label{sec:inference}
Typical ways to estimate the parameters include Bayesian inference and
maximum likelihood estimation (MLE). Bayesian inference requires the
posterior sampling for the parameters (see
Appx.~\ref{sec:appen-full-bayesian} for more details), but in our case it
becomes too slow to be feasible if we have many parameters. For more
efficient inference, we adopt Monte Carlo expectation-maximization (MCEM)
\citep{wei1990monte} as an indirect way to maximize the marginal likelihood.
Different from the ordinary EM, the expected value of the log-likelihood
is approximated by a Monte Carlo method in the expectation steps. Posterior
samples of the hidden TPPs are required for the estimation of the expected
values.

The posterior sampling for a hidden TPP is not trivial as it involves the
sampling for an unbounded number of variables and the posterior
stochastic processes of TPPs are not TPPs anymore. Instead, the posterior
stochastic processes of TPPs are spatial point processes (SPPs). Unlike
TPPs, SPPs do not have a natural ordering in space to define a
filtering, hence the CIFs are not well-defined. In order to better describe
a SPP, we need to provide a different type of conditional intensity
function for an SPP:
\begin{definition}
[Papangelou conditional intensity function
	\citep{papangelou1974conditional}] The {\em Papangelou conditional intensity
	function} (PCIF) is defined as
\begin{equation*}
    \lambda_{\mathcal{P}}(t)=\lim_{\Delta t \rightarrow 0}\frac{\text{\em Pr}(\text{\em One event in }B_{\Delta t}(t)\ \mid\ [\mathbf{N}\backslash B_{\Delta t}(t)])}{\Delta t},
\end{equation*}
where \(B_{\Delta t}(t)=[t,t+\Delta t]\),
\(\mathbf{N}=[\mathbf{Z},\mathbf{x}]\) is the whole set of hidden TPPs and
the evidence, and \([\mathbf{N}\backslash B_{\Delta t}(t)]\) is the
information of the point processes \(\mathbf{N}\) outside \(B_{\Delta
t}(t)\).
\end{definition}
Different from a CIF, which is only conditional on the history, a PCIF is
conditional on the whole space. For a Poisson process, PCIF is equivalent
to CIF as the probability to have an event at any infinitesimal interval is
independent of the time or location.

The PCIF for our model is given in Prop.~\ref{prop:posterior-not-poisson}. See Appx.~\ref{sec:appen-posterior-pdf} for more details.
\begin{proposition}
\label{prop:posterior-not-poisson}
The PCIF for the posterior point process of \(Z_{\ell,k}\) is
\begin{align}
    \lambda_{\mathcal{P};\ell,k}(t) 
    =&\lambda_{\ell,k}(t) \prod_{i=1}^{K_{\ell-1}}\Bigg(\!\!\exp{\left(-\Phi_{(\ell,k) \rightarrow (\ell-1,i)}(T-t)\right)} \nonumber\\
    &\qquad\qquad \prod_{t_{\ell-1,i,j}>t}\!\!\!\frac{\lambda'_{\ell-1,i}(t_{\ell-1,i,j},t)}{\lambda_{\ell-1,i}(t_{\ell-1,i,j})}\Bigg), \label{eq:posterior-pcif}
\end{align}
where \(\lambda'_{\ell-1,i}(x,t)=\lambda_{\ell-1,i}(x)+\phi_{(\ell,k) \rightarrow (\ell-1,i)}(x-t)\) and \(\Phi_{(\ell,k) \rightarrow (\ell-1,i)}(x)=\int_0^{x}\phi_{\bm{\theta}_{(\ell,k) \rightarrow (\ell-1,i)}}(\tau)d\tau\).
\end{proposition}
From Prop.~\ref{prop:posterior-not-poisson}, we can see that the PCIF at time \(t\) is not only controlled by the events on layers
\(\ell+1\) through \(\lambda_{\ell,k}(\cdot)\), but controlled by the events on layers \(\ell\) and \(\ell-1\) through \(\{\lambda_{\ell-1,i}(\cdot)\}_{i=1}^{K_{\ell-1}}\).

\subsection{MCMC} \label{sec:mcmc}

As the PCIF has complex correlations both within and between nodes, it is
hard to directly do posterior sampling for the hidden TPPs.  Considering the
simplest case, if we only have one hidden layer, our model just becomes a
multivariate N-SP. In the SPPs
community, SB\&D is more popular than M-H possibly due to its simplicity and
efficiency
\citep{geyer1994simulation,clifford1994comparison,moller2003statistical}.
Unfortunately, the sufficient condition for the convergence of SB\&D is not
satisfied in our case, and it is an open question of whether SB\&D converges
for our model.  We provide
more details in Appx.~\ref{appen:sbd}. Thus, we
devised a different Markov chain equipped with auxiliary variables that converges quickly to the true posterior distribution.
Compared with a naive MCMC sampler which just re-samples all the hidden events from homogeneous Poisson processes every time as
the proposal, our posterior sampling is much more efficient
due to the help of auxiliary variables.
\begin{remark}
Our MCMC can be applied to non-casual kernels (i.e., the kernels can be functions which have non-zero values in the whole space, like a Gaussian function) and SPPs with any dimensions, not only temporal DN-SPs. No matter how we choose the
kernel, the detailed balance and ergodicity conditions are still satisfied,
and thus the MCMC sampler still converges to the posterior distribution.
\end{remark}
\subsubsection{Virtual Events}
Similar to prior work
\citep{rao2011fast,rao2013fast,qin2015auxiliary,shelton2018hawkes}, we add
virtual events as auxiliary variables for our MCMC sampler. They work by
providing
the candidates for the real events of the hidden TPPs, and they do not
contribute to any intensity functions. With the help of the virtual events,
we only need to search for the real events where the virtual events appear,
instead of the whole space.  We explore all possible real event locations
by resampling virtual events.

For each data point \(\mathbf{x}\), there are \(L\) layers of virtual TPPs
(VPPs)
\(\tilde{\mathbf{Z}}=\{\tilde{\mathbf{Z}}_{1},\dots,\tilde{\mathbf{Z}}_{L}\}\)
aligning with the hidden real point processes (RPPs) \(\mathbf{Z}\),
where \(\tilde{\mathbf{Z}}_{\ell}=\{\tilde{Z}_{\ell,k}\}_{k=1}^{K_\ell}\)
with \(\tilde{Z}_{\ell,k}\) as a virtual TPP. The CIF for
\(\tilde{Z}_{\ell,k}\) conditioned on \(\mathbf{Z}_{\ell-1}\) is 
\begin{equation}
	\tilde{\lambda}_{\ell,k}(\tilde{t}) = 
	\tilde{\mu}_{\ell,k}+\!\sum_{i=1}^{K_{\ell-1}}\!\sum_{t_{\ell-1,i,j}}\!\!\tilde{\phi}_{\tilde{\bm{\theta}}_{(\ell-1,i) \rightarrow (\ell,k)}}(t_{\ell-1,i,j}-\tilde{t}) \label{eq:virtual-intensity} 
\end{equation}
where \(\tilde{\mu}_{\ell,k} \geq 0\) is the base rate,
\(\tilde{\phi}_{\tilde{\bm{\theta}}_{(\ell-1,i) \rightarrow
(\ell,k)}}(\cdot)\) is the virtual kernel function, which we assume is a
gamma kernel. Note that
\(\tilde{\phi}_{\tilde{\bm{\theta}}_{(\ell-1,i) \rightarrow
(\ell,k)}}(t_{\ell-1,i,j}-\tilde{t})\) evolves in the \textbf{opposite direction} to
\(\phi_{\bm{\theta}_{(\ell+1,i) \rightarrow (\ell,k)}}(t-t_{\ell+1,i,j})\),
which we will rationalize later.  However, the intensity of
virtual events at layer \(\ell\) depends on the \textbf{real events} at
layer \(\ell-1\), not the virtual events there.

\begin{figure*}[tb]
\centering
\begin{subfigure}[b]{0.22\textwidth}
\begin{adjustbox}{width=\textwidth}
\begin{tikzpicture}
[
recnode2/.style={      rectangle,
      thick,
      text width=5cm,
      align=right,
      minimum height=1.25cm},   
real_event_line/.style={red,-{Circle[length=3mm,width=3mm]},line width=1mm},
virtual_event_line/.style={blue,-{Rays[length=3mm,width=3mm]},line width=1mm},
selectnode/.style={ellipse, minimum height=1.2cm, minimum width=6mm, fill=black!20},
selectline/.style={-{Latex[length=2mm,width=4mm]}},
connectline/.style={-Stealth,line width=1.5mm},
timeline/.style={-Triangle, line width=1mm},
]
\node[recnode2] (1) {};
\node[recnode2,below=1.5cm of 1.south east, anchor=north east] (2) {};
\draw[timeline] (1.west) -- (1.east);
\foreach \i in {1mm,10mm,15mm,30mm}
\draw[real_event_line] ($(1.west)!\i!(1.east)$) -- ($(1.west)!\i!(1.east)+(0,1)$);
\foreach \i in {5mm,19mm,24mm,34mm}
{
\node[selectnode] at ($(1.west)!\i!(1.east)+(0,0.5)$) {};
\draw[virtual_event_line] ($(1.west)!\i!(1.east)$) -- ($(1.west)!\i!(1.east)+(0,1)$);
\draw[selectline] ($(1.west)!\i!(1.east)+(0,1.5)$) -- ($(1.west)!\i!(1.east)+(0,1.1)$);
}
\draw[connectline] ($(1.south)$) -- ($(2.north)+(0,0.5)$);
\foreach \i in {1mm,10mm,15mm,30mm}
\draw[real_event_line] ($(2.south west)!\i!(2.south east)$) -- ($(2.south west)!\i!(2.south east)+(0,1)$);
\draw[timeline] (2.south west) -- (2.south east);
\foreach \i in {26mm,38mm}{
\node[selectnode] at ($(2.south west)!\i!(2.south east)+(0,0.5)$) {};
\draw[virtual_event_line] ($(2.south west)!\i!(2.south east)$) -- ($(2.south west)!\i!(2.south east)+(0,1)$);
\draw[selectline] ($(2.south west)!\i!(2.south east)+(0,1.5)$) -- ($(2.south west)!\i!(2.south east)+(0,1.1)$);
}
\end{tikzpicture}
\end{adjustbox}
\caption{Re-sample}
\end{subfigure}
~
\begin{subfigure}[b]{0.22\textwidth}
\begin{adjustbox}{width=\textwidth}
\begin{tikzpicture}
[
recnode2/.style={      rectangle,
      thick,
      text width=5cm,
      align=right,
      minimum height=1.25cm},   
real_event_line/.style={red,-{Circle[length=3mm,width=3mm]},line width=1mm},
virtual_event_line/.style={blue,-{Rays[length=3mm,width=3mm]},line width=1mm},
selectnode/.style={ellipse, minimum height=1.2cm, minimum width=6mm, fill=black!20},
selectline/.style={-{Latex[length=2mm,width=4mm]}},
connectline/.style={-Stealth,line width=1.5mm},
timeline/.style={-Triangle, line width=1mm},
]
\node[recnode2] (1) {};
\node[recnode2,below=1.5cm of 1.south east, anchor=north east] (2) {};
\draw[timeline] (1.west) -- (1.east);
\foreach \i in {1mm,15mm,30mm}
\draw[real_event_line] ($(1.west)!\i!(1.east)$) -- ($(1.west)!\i!(1.east)+(0,1)$);
\foreach \i in {5mm,19mm,24mm,34mm}{
\draw[virtual_event_line] ($(1.west)!\i!(1.east)$) -- ($(1.west)!\i!(1.east)+(0,1)$);
}
\draw[connectline] ($(1.south)$) -- ($(2.north)+(0,0.5)$);
\foreach \i in {1mm,15mm,30mm}
\draw[real_event_line] ($(2.south west)!\i!(2.south east)$) -- ($(2.south west)!\i!(2.south east)+(0,1)$);
\draw[timeline] (2.south west) -- (2.south east);
\foreach \i in {5mm,10mm,19mm,24mm,34mm}
\draw[virtual_event_line] ($(2.south west)!\i!(2.south east)$) -- ($(2.south west)!\i!(2.south east)+(0,1)$);
\node[selectnode] at ($(1.west)+(10mm,0)+(0,0.5)$) {};
\draw[real_event_line] ($(1.west)+(10mm,0)$) -- ($(1.west)+(10mm,0)+(0,1)$);
\draw[selectline] ($(1.west)+(10mm,1.5)$) -- ($(1.west)+(10mm,1.1)$);
\node[selectnode] at ($(2.south west)+(10mm,0)+(0,0.5)$) {};
\draw[virtual_event_line] ($(2.south west)+(10mm,0)$) -- ($(2.south west)+(10mm,0)+(0,1)$);
\draw[selectline] ($(2.south west)+(10mm,1.5)$) -- ($(2.south west)+(10mm,1.1)$);
\end{tikzpicture}
\end{adjustbox}
\caption{Flip: Real to Virtual}
\end{subfigure}
~
\begin{subfigure}[b]{0.22\textwidth}
\begin{adjustbox}{width=\textwidth}
\begin{tikzpicture}
[
recnode2/.style={      rectangle,
      thick,
      text width=5cm,
      align=right,
      minimum height=1.25cm},   
real_event_line/.style={red,-{Circle[length=3mm,width=3mm]},line width=1mm},
virtual_event_line/.style={blue,-{Rays[length=3mm,width=3mm]},line width=1mm},
selectnode/.style={ellipse, minimum height=1.2cm, minimum width=6mm, fill=black!20},
selectline/.style={-{Latex[length=2mm,width=4mm]}},
connectline/.style={-Stealth,line width=1.5mm},
timeline/.style={-Triangle, line width=1mm},
]
\node[recnode2] (1) {};
\node[recnode2,below=1.5cm of 1.south east, anchor=north east] (2) {};
\draw[timeline] (1.west) -- (1.east);
\foreach \i in {1mm,10mm,15mm,30mm}
\draw[real_event_line] ($(1.west)!\i!(1.east)$) -- ($(1.west)!\i!(1.east)+(0,1)$);
\foreach \i in {5mm,24mm,34mm}
\draw[virtual_event_line] ($(1.west)!\i!(1.east)$) -- ($(1.west)!\i!(1.east)+(0,1)$);
\draw[connectline] ($(1.south)$) -- ($(2.north)+(0,0.5)$);
\foreach \i in {1mm,10mm,15mm,30mm}
\draw[real_event_line] ($(2.south west)!\i!(2.south east)$) -- ($(2.south west)!\i!(2.south east)+(0,1)$);
\draw[timeline] (2.south west) -- (2.south east);
\foreach \i in {5mm,24mm,34mm}
\draw[virtual_event_line] ($(2.south west)!\i!(2.south east)$) -- ($(2.south west)!\i!(2.south east)+(0,1)$);
\node[selectnode] at ($(1.west)+(19mm,0)+(0,0.5)$) {};
\draw[virtual_event_line] ($(1.west)+(19mm,0)$) -- ($(1.west)+(19mm,0)+(0,1)$);
\draw[selectline] ($(1.west)+(19mm,1.5)$) -- ($(1.west)+(19mm,1.1)$);
\node[selectnode] at ($(2.south west)+(19mm,0)+(0,0.5)$) {};
\draw[real_event_line] ($(2.south west)+(19mm,0)$) -- ($(2.south west)+(19mm,0)+(0,1)$);
\draw[selectline] ($(2.south west)+(19mm,1.5)$) -- ($(2.south west)+(19mm,1.1)$);
\end{tikzpicture}
\end{adjustbox}
\caption{Flip: Virtual to Real}
\end{subfigure}
~
\begin{subfigure}[b]{0.22\textwidth}
\begin{adjustbox}{width=\textwidth}
\begin{tikzpicture}
[
recnode2/.style={      rectangle,
      thick,
      text width=5cm,
      align=right,
      minimum height=1.25cm},   
real_event_line/.style={red,-{Circle[length=3mm,width=3mm]},line width=1mm},
virtual_event_line/.style={blue,-{Rays[length=3mm,width=3mm]},line width=1mm},
selectnode/.style={ellipse, minimum height=1.2cm, minimum width=6mm, fill=black!20},
selectline/.style={-{Latex[length=2mm,width=4mm]}},
connectline/.style={-Stealth,line width=1.5mm},
timeline/.style={-Triangle, line width=1mm},
]
\node[recnode2] (1) {};
\node[recnode2,below=1.5cm of 1.south east, anchor=north east] (2) {};
\draw[timeline] (1.west) -- (1.east);
\foreach \i in {1mm,15mm,30mm}
\draw[real_event_line] ($(1.west)!\i!(1.east)$) -- ($(1.west)!\i!(1.east)+(0,1)$);
\foreach \i in {5mm,19mm,24mm}
\draw[virtual_event_line] ($(1.west)!\i!(1.east)$) -- ($(1.west)!\i!(1.east)+(0,1)$);
\draw[connectline] ($(1.south)$) -- ($(2.north)+(0,0.5)$);
\foreach \i in {1mm,15mm,30mm,34mm}
\draw[real_event_line] ($(2.south west)!\i!(2.south east)$) -- ($(2.south west)!\i!(2.south east)+(0,1)$);
\draw[timeline] (2.south west) -- (2.south east);
\foreach \i in {5mm,10mm,19mm,24mm}
\draw[virtual_event_line] ($(2.south west)!\i!(2.south east)$) -- ($(2.south west)!\i!(2.south east)+(0,1)$);
\node[selectnode] at ($(1.west)+(10mm,0)+(0,0.5)$) {};
\node[selectnode] at ($(1.west)+(34mm,0)+(0,0.5)$) {};
\draw[{Latex[length=2mm,width=4mm]}-{Latex[length=2mm,width=4mm]}] ($(1.west)+(10mm,1.1)$) to [out=30,in=150] ($(1.west)+(34mm,1.1)$);
\draw[real_event_line] ($(1.west)+(10mm,0)$) -- ($(1.west)+(10mm,0)+(0,1)$);
\draw[virtual_event_line] ($(1.west)+(34mm,0)$) -- ($(1.west)+(34mm,0)+(0,1)$);
\node[selectnode] at ($(2.south west)+(10mm,0)+(0,0.5)$) {};
\node[selectnode] at ($(2.south west)+(34mm,0)+(0,0.5)$) {};
\draw[real_event_line] ($(2.south west)+(34mm,0)$) -- ($(2.south west)+(34mm,0)+(0,1)$);
\draw[virtual_event_line] ($(2.south west)+(10mm,0)$) -- ($(2.south west)+(10mm,0)+(0,1)$);
\draw[{Latex[length=2mm,width=4mm]}-{Latex[length=2mm,width=4mm]}] ($(2.south west)+(10mm,1.1)$) to [out=30,in=150] ($(2.south west)+(34mm,1.1)$);
\end{tikzpicture}
\end{adjustbox}
\caption{Swap}
\end{subfigure}
\vspace{.1in}
\caption{Examples For Sampler Moves. \protect\tikz \protect\draw[red,-{Circle[length=2mm,width=2mm]},line width=0.66mm] (0,-0.3) -- (0,0.3);
 Represents a Real Event.\protect\tikz \protect\draw[blue,-{Rays[length=2mm,width=2mm]},line width=0.66mm] (0,-0.3) -- (0,0.3); Represents a Virtual Event.}
\label{fig:sampler-moves}
\end{figure*}

\subsubsection{Complete Likelihood}
The complete likelihood for the joint RPPs and VPPs for a data point is 
\begin{equation}
    p(\mathbf{x}, \mathbf{Z}\!\!=\!\!\mathbf{z}, \tilde{\mathbf{Z}}\!\!=\!\!\tilde{\mathbf{z}})
    =p(\mathbf{z}_{L}) \prod_{\ell=0}^{L-1}p(\mathbf{z}_{\ell}\mid\mathbf{z}_{\ell+1}) \tilde{p}(\tilde{\mathbf{z}}_{\ell+1} \mid\mathbf{z}_{\ell}), \label{eq:complete-likelihood}
\end{equation}
where
\begin{align*}
    p(\mathbf{z}_{L})&=\textstyle\prod_{k=1}^{K_L}p(\mathbf{z}_{L,k}),\\
	p(\mathbf{z}_{\ell}\mid\mathbf{z}_{\ell+1})&=\textstyle\prod_{k=1}^{K_\ell}p(\mathbf{z}_{\ell,k}\mid\mathbf{z}_{\ell+1}),\\
	\tilde{p}(\tilde{\mathbf{z}}_{\ell}\mid\mathbf{z}_{\ell-1})&=\textstyle\prod_{k=1}^{K_{\ell}}\tilde{p}(\tilde{\mathbf{z}}_{\ell,k}
	\mid\mathbf{z}_{\ell-1})\,\,.
\end{align*}
The likelihood of \(\mathbf{z}_{L,k}\) is
\begin{equation*}
    p(\mathbf{z}_{L,k})=\exp\left(-\mu_{L,k}T\right)\mu_{L,k}^{m_{L,k}},
\end{equation*} where \(m_{L,k}\) is the number of events drawn from \(Z_{L,k}\).

The likelihood of \(\mathbf{z}_{\ell,k}\) for \(0\leq \ell \leq L-1\) is
\begin{equation*}
	p(\mathbf{z}_{\ell,k}|\mathbf{z}_{\ell+1})\!=\!\exp\!\Bigg(\ \ \;\smashoperator{\sum_{t_{\ell,k,j}\leq T}} \log \lambda_{\ell,k}(t_{\ell,k,j})
    -\!\int_0^T\!\!\!\!\lambda_{\ell,k}(t)\,dt\Bigg)\,.
\end{equation*}
(Here and for the rest of the paper, we let
 \(\mathbf{z}_{0,k}\) be \(\mathbf{x}_{k}\).)
 
The likelihood of \(\tilde{\mathbf{z}}_{\ell,k}\) for \(1\leq \ell\leq L\) is
\begin{equation*}
    \tilde{p}(\tilde{\mathbf{z}}_{\ell,k} |\mathbf{z}_{\ell-1})
	\!=\!\exp\!\Bigg(\ \ \;\smashoperator{\sum_{\tilde{t}_{\ell,k,j}\leq T}} \log \tilde{\lambda}_{\ell,k}(\tilde{t}_{\ell,k,j})
    -\!\int_0^T\!\!\!\!\tilde{\lambda}_{\ell,k}(t)\,dt\Bigg)\,.
\end{equation*}

\subsubsection{Sampler Moves}
During the sampling process, we first select a hidden TPP uniformly and then apply a move selected randomly from the following three types with a predetermined probability distribution. See Fig.~\ref{fig:sampler-moves} for illustration. More details can be found in Appx.~\ref{sec:sampler-moves}. 

\textit{Move 1: Re-sample virtual events.} This move re-samples the virtual
events for a VPP. The dimensionality of the sample space is changed after each
re-sampling. But the determinant of the Jacobian matrix, introduced as the
correction for the changes of variables in reversible-jump MCMC
\citep{green1995reversible}, is \(1\), since the new variables are independent of the current virtual events of the Markov chain. The acceptance probability is always \(1\) for this move. 

\textit{Move 2: Flip.} We uniformly pick an event from the union of the samples from the RPP and the VPP, and then propose to change the type for that event. If the type of the picked event is real, we propose to flip it to be a virtual event, and vice versa.

\textit{Move 3: Swap.} One event is picked uniformly from the samples for
each of the RPP and the VPP. Then we propose to swap the types of these two events, \textit{i.e.}, the real event becomes a virtual event and vice versa. 

Suppose the proposals for the changes are to adjust the events in
\(\mathbf{z}_{\ell,k}\) and \(\tilde{\mathbf{z}}_{\ell,k}\) to become the events in \(\mathbf{z}'_{\ell,k}\) and  \(\tilde{\mathbf{z}}'_{\ell,k}\).
Then the likelihood ratio is
\begin{align*}
	\mathcal{P} 
	&=\frac{p(\mathbf{z}'_{\ell,k}|\mathbf{z}_{\ell+1})
	\tilde{p}(\tilde{\mathbf{z}}'_{\ell,k}|\mathbf{z}_{\ell-1})}{p(\mathbf{z}_{\ell,k}|\mathbf{z}_{\ell+1}) \tilde{p}(\tilde{\mathbf{z}}_{\ell,k}|\mathbf{z}_{\ell-1})}
	\cdot
\frac{ p(\mathbf{z}_{\ell-1}|\mathbf{z}'_{\ell}) \tilde{p}(\tilde{\mathbf{z}}_{\ell+1}|\mathbf{z}'_{\ell})}{
	p(\mathbf{z}_{\ell-1}|\mathbf{z}_{\ell}) \tilde{p}(\tilde{\mathbf{z}}_{\ell+1}|\mathbf{z}_{\ell})},
\end{align*}
where
	\(
  p(\mathbf{z}_{\ell-1}\mid\mathbf{z}_{\ell})=\textstyle\prod_{k=1}^{K_{\ell-1}}p(\mathbf{z}_{\ell-1,k}\mid\mathbf{z}_{\ell}),\ 
	\text{and}\ 
  \tilde{p}(\tilde{\mathbf{z}}_{\ell+1}\mid\mathbf{z}_{\ell})=\textstyle\prod_{k=1}^{K_{\ell+1}}\tilde{p}(\tilde{\mathbf{z}}_{\ell+1,k}\mid\mathbf{z}_{\ell}).
	\)

The ratio for the proposal probability is \(1\) for both \textit{Move 2} and \textit{Move 3}.
So the acceptance probability for \textit{Move 2} and \textit{Move 3} is
 \(\min(1, \mathcal{P}\cdot 1)\).
%

It is necessary to have \textit{Move 3} to help accelerate mixing even
though \textit{Move 2} seems to already include the ability to swap through
two consecutive flips. However, often there is a real event that has large
positive contribution to the likelihood in
Eq.~\ref{eq:complete-likelihood}. If we propose to flip this event to a
virtual event, the likelihood ratio \(\mathcal{P}\) would be very small,
hence this real event would stay in a same place for a long time.
Similarly, flipping a virtual event first would be unlikely.

With these three sampler moves, we address the form of the VPP intensity functions in Eq.~\ref{eq:virtual-intensity}:
\begin{compactenum}
    \item The intensity functions for the VPPs are only controlled by the RPPs on the lower layers.
	    This allows us to sample the VPPs directly and efficiently, using the inversion method
	    \citep{cinlar2013introduction} in \textit{Move 1}, to push the information ``upwards.''
    \item The virtual events tend to appear more at the places where the probability is high to have a
	    real event.
	    In particular, events at higher levels tend to proceed those at lower levels.
	    So it is natural to have virtual kernel functions that evolve with time in the reverse
	    direction to the real kernel functions.  The true posterior of the upper level (conditioned on
	    the lower level) is not so simple, but this is a good approximation for the proposal.
    \item The base rates \(\tilde{\mu}_{\ell,k}\) help accelerate the
	    mixing for the point processes not directly connected to the
		evidence.
		Without them, an upper layer can be ``starved'' for virtual events (necessary to allow the
		addition and movement of real points) if a lower layer has few real events.
\end{compactenum}

\subsubsection{Update Virtual Kernels Parameters}
Because we want the distribution of the proposed virtual events to be as close as possible to the posterior distribution of the real events, we would like to maximize the likelihood of the parameters for the VPPs assuming the posterior samples for the real events are drawn from the VPPs.
That is, we adjust the sampler to make it more efficient by tuning the parameters of the VPPs (as the sampler is valid across different VPP
parameters).

As the number of events usually varies significantly for different evidence samples, we assume the base rates \(\tilde{\mu}_{n,L,i}\) and
\(\mu_{n,i}\) on the top layer are different for each data point \(\mathbf{x}_n\), where \(n\) represents the data index.

\newcommand\llh{\text{llh}}

Let
\(\mathbf{z}_n^{(1)},\dots,\mathbf{z}_n^{\mathcal{(S)}}\) be the posterior samples from the distribution
\(p(\mathbf{Z}_n\mid\mathbf{x}_n;\bm{\mu}_n,\bm{\theta})\), where \(\bm{\mu}_n=\left[\mu_{n,i}\right]_{i=1}^{K_L}\) are the parameters of
the HPPs on the top layer and \(\bm{\theta}=\left[\bm{\theta}_{(\ell+1,*) \rightarrow (\ell,*)}\right]_{\ell=0}^{L-1}\) are the parameters of the kernel functions.
Given the log-likelihood of the parameters of the VPPs \textit{w.r.t} the real events
\begin{equation*}
	\tilde{\llh}\left(\tilde{\bm{\theta}},\tilde{\bm{\mu}}_n;\mathbf{x}_n,\mathbf{z}_n\right)
	=\sum_{\ell=1}^L\tilde{\llh}_{(\ell-1,*)\rightarrow(\ell,*);\mathbf{x}_n,\mathbf{z}_n}
\end{equation*}
where
\begin{align*}
	&\tilde{\llh}_{{(\ell-1,*) \rightarrow (\ell,*)};\mathbf{x}_n,\mathbf{z}_{n}}\\
=&\sum_{k=1}^{K_\ell}\Bigg(\sum_{t_{n,\ell,k,j}\leq T} \log
	\tilde{\lambda}_{n,\ell,k}(t_{n,\ell,k,j})-\int_0^T\tilde{\lambda}_{n,\ell,k}(t)dt\Bigg)\,\,.
\end{align*}
The update rules are
\begin{align}
    \tilde{\bm{\mu}}_n \leftarrow
	&\tilde{\bm{\mu}}_n+\frac{\tilde{r}}{\mathcal{S}N}\sum_{s=1}^{\mathcal{S}}\nabla_{\tilde{\bm{\mu}}_n}\tilde{\llh}\left(\tilde{\bm{\theta}}, \tilde{\bm{\mu}}_n;\mathbf{x}_n,\mathbf{z}_n^{(s)}\right),
    \label{eq:update-virtual-mu}\\
	\tilde{\bm{\theta}}\leftarrow&\tilde{\bm{\theta}}+\frac{\tilde{r}}{\mathcal{S}N}\sum_{n=1}^N\sum_{s=1}^{\mathcal{S}}\nabla_{\tilde{\bm{\theta}}}\tilde{\llh}\left(\tilde{\bm{\theta}},\tilde{\bm{\mu}}_n;\mathbf{x}_n,\mathbf{z}_n^{(s)}\right),
    \label{eq:update-virtual-kernel}
\end{align}
where\[\tilde{\bm{\mu}}_{n}=\left[\left[\tilde{\mu}_{n,\ell,k}\right]_{\ell=1}^L\right]_{k=1}^{K_\ell},\] 
\[\tilde{\bm{\theta}}=\left[\left[\left[\bm{\tilde{\theta}}_{(\ell-1,i)\rightarrow(\ell,j)}\right]_{i=1}^{K_{\ell-1}}\right]_{j=1}^{K_\ell}\right]_{\ell=1}^L,\]
\(N\) is the number of data points, and \(\tilde{r} > 0\) is the step size for optimizing. See Appx.~\ref{sec:appen-opt-vpps} for the gradient.

\subsection{MCEM}
MCEM iteratively applies the following two-step process until the parameters for the RPPs converge.

\textit{Step 1.} Generate posterior samples \(\mathbf{z}_n^{(1)},\dots,\mathbf{z}_n^{\mathcal{(S)}}\) from the distribution
\(p(\mathbf{Z}_n\mid\mathbf{x}_n;\bm{\mu}_n,\bm{\theta})\). 

\textit{Step 2.} Update the parameters.  Given the log-likelihood function of the parameters of the RPPs \textit{w.r.t} the real events
\begin{align}
	&\llh(\bm{\theta},\bm{\mu}_n;\mathbf{x}_n,\mathbf{z}_n) \nonumber\\
    =&\log\left(\prod_{i=1}^{K_L}p(\mathbf{z}_{n,L,i})\cdot \prod_{\ell=0}^{L-1}\prod_{i=1}^{K_\ell}p(\mathbf{z}_{n,\ell,i}\mid\mathbf{z}_{n,\ell+1})\right), \label{eq:ll-to-optimize}
\end{align}
the update rules for the parameters are
\begin{align}
	\bm{\mu}_n&\leftarrow \argmax_{\bm{\mu}_n}\left\{\frac{1}{\mathcal{S}}\sum_{s=1}^{\mathcal{S}}\llh(\bm{\bm{\theta},\mu}_n;\mathbf{x}_n,\mathbf{z}_n^{(s)})\right\}, \label{eq:update-real-mu}\\
	\bm{\theta}&\leftarrow\bm{\theta}+\frac{r}{\mathcal{S} N}\sum_{n=1}^N\sum_{s=1}^{\mathcal{S}}\nabla_{\bm{\theta}}\llh(\bm{\theta},\bm{\mu}_n;\mathbf{x}_n,\mathbf{z}_n^{(s)}), \label{eq:update-real-kernel}
\end{align}
where \(N\) is the number of data points and \(r > 0\) is the step size for optimizing. See Appx.~\ref{sec:appen-opt-rpps} for the gradient and maximization formula.

Full maximization is utilized for \(\bm{\mu}_n\) in Eq.~\ref{eq:update-real-mu}, and
ascent-based MCEM \citep{caffo2005ascent} is used for \(\bm{\theta}\) in
Eq.~\ref{eq:update-real-kernel}. When the sample size \(\mathcal{S}\) goes
to infinity and we update the parameters as in Eq.~\ref{eq:update-real-kernel}, the expected value of the log-likelihood
in Eq.~\ref{eq:ll-to-optimize} increases at each iteration with probability
converging to 1  \citep{caffo2005ascent}. The general theory for the
convergence of MCEM has not been well-established. Different senses and
different approaches for the convergence analysis are given by \citet{neath2013convergence}. We tried various settings for MCEM and choose the one that is stable and computationally efficient. Adam \citep{DBLP:journals/corr/KingmaB14} was utilized for the optimization of the parameters of the kernel and virtual kernel functions. Notice that we are not changing the parameters for the real or virtual kernels during MCMC sampling. The posterior samples for the real events are collected from the MCMC sampler after 
it reaches convergence. Pseudo-code is given in Appx.~\ref{sec:inf-alg}.

\section{EXPERIMENTS} \label{sec:experiments}
The code is available online at \url{https://github.com/hongchengkuan/Deep-Neyman-Scott-Processes}.
\subsection{Architectures}
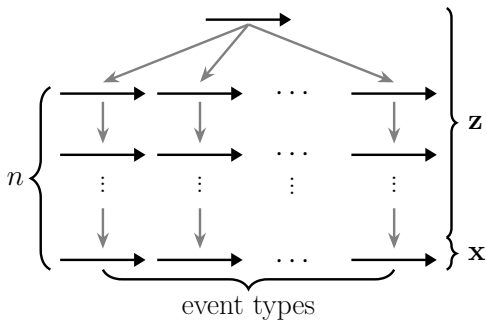
\begin{figure}
\centering
\begin{adjustbox}{width=0.38\textwidth}
\begin{tikzpicture}
[
recnode2/.style={      rectangle,
      thick,
      text width=1.5cm,
      align=center},   
real_event_line/.style={red,-{Circle[length=3mm,width=3mm]},line width=1mm},
virtual_event_line/.style={blue,-{Rays[length=3mm,width=3mm]},line width=1mm},
selectnode/.style={ellipse, minimum height=1.2cm, minimum width=6mm, fill=black!20},
selectline/.style={-{Latex[length=2mm,width=4mm]}},
connectline/.style={gray, -Stealth, line width=0.5mm},
timeline/.style={-Triangle, line width=0.5mm},
]
\node[recnode2] (1) {};
\node[recnode2,below left=1cm and 3cm of 1.south east, anchor=north east] (2) {};
\node[recnode2, below left=1cm and 1cm of 1.south east, anchor=north east] (3) {};
\node[recnode2, below right=1cm and 0.9cm of 1.south east, anchor=north east] (4) {\LARGE \(\cdots\)};
\node[recnode2, below right=1cm and 3cm of 1.south east, anchor=north east] (5) {};
\node[recnode2, below=1cm of 2.south east, anchor=north east] (6) {};
\node[recnode2, below=1cm of 3.south east, anchor=north east] (7) {};
\node[recnode2, below right=2.25cm and 0.9cm of 1.south east, anchor=north east] (8) {\LARGE \(\cdots\)};
\node[recnode2, below=1cm of 5.south east, anchor=north east] (9) {};
\node[recnode2, below=0.1cm of 6.south , anchor=north] (10) {\LARGE \(\vdots\)};
\node[recnode2, below=0.1cm of 7.south east, anchor=north east] (11) {\LARGE \(\vdots\)};
\node[recnode2, below right=2.65cm and 0.9cm of 1.south east, anchor=north east] (12) {\LARGE \(\vdots\)};
\node[recnode2, below=0.1cm of 9.south east, anchor=north east] (13) {\LARGE \(\vdots\)};
\node[recnode2, below=1cm of 10.south east, anchor=north east] (14) {};
\node[recnode2, below=1cm of 11.south east, anchor=north east] (15) {};
\node[recnode2, below right=4.45cm and 0.9cm of 1.south east, anchor=north east] (16) {\LARGE \(\cdots\)};
\node[recnode2, below=1cm of 13.south east, anchor=north east] (17) {};

\draw[timeline] (1.north west) -- (1.north east);
\foreach \i in {2,3,5}
\draw[connectline] ($(1.north)+(0,-0.1)$) -- ($(\i)+(0,0.1)$);
\foreach \x in {2,3,5}{
\tikzmath{int \i;
\i = \x + 4;}
\draw[connectline] ($(\x)+(0,-0.3)$) -- ($(\i)+(0,0.1)$);
}
\foreach \x in {10,11,13}{
\tikzmath{int \i;
\i = \x + 4;}
\draw[connectline] ($(\x)+(0,-0.6)$) -- ($(\i)+(0,0.1)$);
}
\foreach \i in {2,3,5,6,7,9,14,15,17}
\draw[timeline] ($(\i.south west)$) -- ($(\i.south east)$);
\draw[decoration={brace,mirror,raise=0.2cm,amplitude=0.4cm},decorate, line width=0.5mm]
  (14.south) -- node[below=0.6cm] {\LARGE event types} (17.south);
\draw[decoration={brace,mirror,raise=0.2cm,amplitude=0.4cm},decorate, line width=0.5mm]
	(2.west) -- node[left=0.6cm] {\LARGE\(n\)} ($(14.south west) + (0.0,-0.2)$);
\draw[decoration={brace,raise=0.2cm,amplitude=0.2cm},decorate,line width=0.5mm]
	($(17.north east) + (0.0,0.2)$) -- node[right=0.5cm] {\LARGE \(\mathbf{x}\)} ($(17.south east) + (0.0,-0.2)$);
\draw[decoration={brace,raise=0.2cm,amplitude=0.2cm},decorate,line width=0.5mm]
	($(5.north east) + (0.0,1.5)$) -- node[right=0.5cm] {\LARGE \(\mathbf{z}\)} ($(17.north east) + (0.0,0.2)$);
\end{tikzpicture}
\end{adjustbox}
\vspace{.15in}
\caption{\(n\)-hidden}
\label{fig:architectures}
\vskip -0.1in
\end{figure}

We constructed hierarchical models as in Fig.~\ref{fig:architectures}. Black horizontal arrows are point
processes.  Gray arrows are model connections.
For the 1-hidden model, we only have one layer of hidden TPPs and there is only one TPP in total on the top layer. The hidden TPP is connected to all the types of events in the evidence.

For the 2-hidden model, we have the one hidden TPP on layer 1 for each type in the evidence with
a single connection between matched hidden- and observed-TPPs,
\textit{i.e.}, \(\phi_{\bm{\theta}_{(1,i)\rightarrow(0,j)}}(\cdot)=0\) if \(i\neq j\). There is also only one hidden TPP on the top layer, which is connected to all hidden TPPs on layer 1.

The \(n\)-hidden model can be constructed by adding more hidden layers for
each type similarly. More details about training, testing, and prediction are given in Appx.~\ref{sec:exp-details}. 
For each dataset, the \underline{\textbf{best}} result is shown in bold and underlined, the \textbf{runner-up} in bold.
\subsection{Synthetic Data Experiments}
We use synthetic data to illustrate the modeling power of multiple layers.
The synthetic data are generated from DN-SPs
with differing numbers of hidden layers and 2 event types.  They are divided into
training and test sets. Then we apply our models with different number of
hidden layers to train and test the synthetic datasets. The log-likelihood
per event, time prediction root mean squared error (RMSE) and type
prediction accuracy for the test set are shown in Table
\ref{table:synthetic-results}. The leftmost column represents the different
depths we use to generate synthetic data and the right 3 columns are the
results when we use different model depths for training and testing. For log-likelihood
(which our model is trained for), increasing depth helps, particularly up
to the depth of the data. Accuracy and RMSE have similar behavior.

\newcommand{\tikzmark}[2]{\tikz[overlay,remember picture,
  baseline=(#1.base)] \node (#1) {#2};}

\begin{table*}[hbt]
\caption{Synthetic Experimental Results}
\label{table:synthetic-results}
\begin{center}
\begin{small}
\begin{sc}
\begin{tabular}{lccrd{1.3}}
\toprule
Synthetic Datasets & Model & \multicolumn{1}{c}{Log-likelihood} & RMSE & \multicolumn{1}{c}{Accuracy}\\
\midrule
\multirow{5}{12em}{2-hidden} &1-hidden    &\multicolumn{1}{d{1.3}}{-0.006} &
	\multicolumn{1}{d{1.3}}{1.048} & 0.722 \\ [0.5ex]

	&\multicolumn{1}{c}{\tikzmark{leftA}{2-hidden}} & \multicolumn{1}{d{1.3}}{0.286} &
	\multicolumn{1}{U{.}{.}{1.3}}{0.942} & \multicolumn{1}{Z{.}{.}{1.3}}{0.760\tikzmark{rightA}{}}\\ [0.5ex]
&3-hidden    & \multicolumn{1}{Z{.}{.}{1.3}}{0.301} & \multicolumn{1}{Z{.}{.}{1.3}}{1.010} &
	\multicolumn{1}{U{.}{.}{1.3}}{0.762} \\ [0.5ex]
&4-hidden    & \multicolumn{1}{U{.}{.}{1.3}}{0.304}  & \multicolumn{1}{d{1.3}}{1.189} & 0.757         \\
\midrule
\multirow{5}{12em}{3-hidden} &1-hidden    &\multicolumn{1}{d{1.3}}{0.528} &
	\multicolumn{1}{d{1.3}}{0.731} & 0.782 \\ [0.5ex]
&2-hidden & \multicolumn{1}{Z{.}{.}{1.3}}{0.822} & \multicolumn{1}{U{.}{.}{1.3}}{0.611} &
	\multicolumn{1}{Z{.}{.}{1.3}}{0.812}\\  [0.5ex]
	&\multicolumn{1}{c}{\tikzmark{leftB}{3-hidden}}    & \multicolumn{1}{U{.}{.}{1.3}}{0.835} &
	\multicolumn{1}{Z{.}{.}{1.3}}{0.613} & \multicolumn{1}{U{.}{.}{1.3}}{0.814\tikzmark{rightB}{}} \\[0.5ex]
	&4-hidden    & \multicolumn{1}{d{1.3}}{0.831} & \multicolumn{1}{d{1.3}}{0.670} & 0.809         \\
\midrule
\multirow{5}{12em}{4-hidden} &1-hidden    &\multicolumn{1}{d{1.3}}{1.177} &
	\multicolumn{1}{d{1.3}}{0.411} & 0.820 \\ [0.5ex]
&2-hidden & \multicolumn{1}{d{1.3}}{1.458} & \multicolumn{1}{d{1.3}}{0.409} &
	\multicolumn{1}{d{1.3}}{0.845}\\ [0.5ex]
&3-hidden    & \multicolumn{1}{Z{.}{.}{1.3}}{1.464} & \multicolumn{1}{U{.}{.}{1.3}}{0.391} &
	\multicolumn{1}{U{.}{.}{1.3}}{0.846} \\  [0.5ex]
	&\multicolumn{1}{c}{\tikzmark{leftC}{4-hidden}}    & \multicolumn{1}{U{.}{.}{1.3}}{1.466} &
	\multicolumn{1}{U{.}{.}{1.3}}{0.391} &    \multicolumn{1}{U{.}{.}{1.3}}{0.846\tikzmark{rightC}{}}      \\
\bottomrule
\end{tabular}
\end{sc}
\end{small}
\end{center}
	\tikz[overlay,remember picture]{%
	\node[draw=black,rounded corners = 0.5ex,fit=(leftA.north west) (rightA.south east),inner sep=-1pt] {};%
	\node[draw=black,rounded corners = 0.5ex,fit=(leftB.north west) (rightB.south east),inner sep=-1pt] {};%
	\node[draw=black,rounded corners = 0.5ex,fit=(leftC.north west) (rightC.south east),inner sep=-1pt] {};%
	}
\vskip -0.1in
\end{table*}


\subsection{Real-world Data Experiments}
We compare our DN-SP to four currently popular continuous-time event modeling methods: MTPP \citep{lian2015multitask}, the neural Hawkes process (NHP) \citep{mei2017neural}, the
self-attentive Hawkes process (SAHP) \citep{zhang2020self}, and the
transformer Hawkes process (THP) \citep{zuo2020transformer}. MTPP combines Gaussian processes and piecewise-constant intensity models \citep{gunawardana2011model}, while the other baselines are based on neural networks. The datasets we
use are retweets, earthquakes, and homicides. The details for these
datasets can be found in Appx.~\ref{sec:datasets-stats}.  Each dataset is
split into training, validation, and test sets randomly five times. We report the mean values 
with the standard deviations shown in parentheses.  Validation is
used to tune the hyperparameters and early stopping for all the datasets for NHP, SAHP and THP. For DN-SPs, validation is only used for early
stopping for retweets dataset, as we only use mini-batch gradient ascent for retweets. Batch gradient ascent and termination based on training are used for earthquakes and homicides.  For MTPP, each task corresponds to a type in our experiments. The model parameters of MTPP are fixed across different sequences and each sequence has their own variational parameters. We use forward sampling to do the prediction for MTPP in a similar way as in Appx.~\ref{sec:pred-details}. 

\begin{table*}[hbt]
\caption{Real-world Experimental Results}
\label{table:results}
\begin{center}
\begin{small}
\begin{sc}
\begin{tabular}{lccrd{1.2}}
\toprule
Datasets & Model & \multicolumn{1}{c}{Log-likelihood} & RMSE(\(\times10^4\)) & \multicolumn{1}{c}{Accuracy}\\
\midrule
\multirow{5}{12em}{Retweets} & MTPP &\multicolumn{1}{d{3.7}}{-13.44(0.18)} & \multicolumn{1}{d{1.7}}{1.64(0.03)} & \multicolumn{1}{d{1.7}}{0.36(0.00)}\\ &NHP    &\multicolumn{1}{d{3.7}}{-6.12(0.03)} & \multicolumn{1}{d{1.7}}{1.64(0.03)} & \multicolumn{1}{d{1.7}}{0.48(0.01)} \\
&SAHP & \multicolumn{1}{d{3.7}}{-4.38(0.17)} & \multicolumn{1}{Z{.}{.}{1.7}}{1.60(0.04)} & \multicolumn{1}{d{1.7}}{0.51(0.03)}\\
&THP    & \multicolumn{1}{d{3.7}}{-4.63 (0.03)} & \multicolumn{1}{U{.}{.}{1.7}}{1.54(0.03)} & \multicolumn{1}{U{.}{.}{1.7}}{0.61(0.00)} \\ [0.5ex]
&1-hidden    & \multicolumn{1}{Z{.}{.}{3.7}}{-4.15(0.07)}  & \multicolumn{1}{Z{.}{.}{1.7}}{1.60(0.03)} & \multicolumn{1}{d{1.7}}{0.48(0.00)}         \\
&2-hidden     & \multicolumn{1}{U{.}{.}{3.7}}{-3.54(0.15)} & \multicolumn{1}{d{1.7}}{1.69(0.07)} &  \multicolumn{1}{Z{.}{.}{1.7}}{0.57(0.00)}\\
\midrule
\multirow{5}{12em}{Earthquakes} & MTPP &\multicolumn{1}{d{3.7}}{-10.45(0.09)} & \multicolumn{1}{d{1.7}}{0.20(0.01)} & \multicolumn{1}{d{1.7}}{0.52(0.00)}
\\ &NHP    &\multicolumn{1}{d{3.7}}{-8.70(0.08)} & \multicolumn{1}{d{1.7}}{0.20(0.01)} & \multicolumn{1}{d{1.7}}{0.39(0.01)} \\
&SAHP & \multicolumn{1}{Z{.}{.}{3.7}}{-8.29(0.25)} & \multicolumn{1}{Z{.}{.}{1.7}}{0.16(0.01)} & \multicolumn{1}{d{1.7}}{0.56(0.03)}\\
&THP    & \multicolumn{1}{U{.}{.}{3.7}}{-8.13(0.05)} & \multicolumn{1}{d{1.7}}{0.20(0.01)} & \multicolumn{1}{U{.}{.}{1.7}}{0.61(0.01)} \\ [0.5ex]
&1-hidden    & \multicolumn{1}{d{3.7}}{-8.45(0.05)} & \multicolumn{1}{U{.}{.}{1.7}}{0.15(0.01)} & \multicolumn{1}{U{.}{.}{1.7}}{0.61(0.01)}         \\
&2-hidden     & \multicolumn{1}{d{3.7}}{-8.43(0.06)} & \multicolumn{1}{Z{.}{.}{1.7}}{0.16(0.01)} &  \multicolumn{1}{Z{.}{.}{1.7}}{0.60(0.01)}\\
\midrule
\multirow{5}{12em}{Homicides} &MTPP    &\multicolumn{1}{d{4.7}}{-18.15(1.40)} & \multicolumn{1}{d{2.7}}{23.14(3.41)} & \multicolumn{1}{d{1.7}}{0.20(0.02)} \\
&NHP    &\multicolumn{1}{d{4.7}}{-21.93(0.93)} & \multicolumn{1}{d{2.7}}{23.13(3.42)} & \multicolumn{1}{d{1.7}}{0.18(0.03)} \\
&SAHP & \multicolumn{1}{d{4.7}}{-14.02(0.33)} & \multicolumn{1}{d{2.7}}{23.13(3.42)} & \multicolumn{1}{d{1.7}}{0.19(0.02)}\\
&THP    & \multicolumn{1}{d{4.7}}{-14.87(0.63)} & \multicolumn{1}{d{2.7}}{23.13(3.42)} & \multicolumn{1}{U{.}{.}{1.7}}{0.26(0.03)} \\ [0.5ex]
&1-hidden    & \multicolumn{1}{Z{.}{.}{4.7}}{-11.68(0.20)} & \multicolumn{1}{U{.}{.}{2.7}}{17.40(2.68)} & \multicolumn{1}{Z{.}{.}{1.7}}{0.25(0.01)}\\
&2-hidden     & \multicolumn{1}{U{.}{.}{4.7}}{-10.78(0.11)} & \multicolumn{1}{Z{.}{.}{2.7}}{17.73(2.84)} &  \multicolumn{1}{Z{.}{.}{1.7}}{0.25(0.02)}\\

\bottomrule
\end{tabular}
\end{sc}
\end{small}
\end{center}
\vskip -0.1in
\end{table*}

The results in Table \ref{table:results} indicate that our model achieves 
competitive results compared to the baselines. We have the best RMSE for
earthquakes and homicides, the best accuracy for earthquakes, and the best likelihood for retweets and homicides.
The likelihood for the earthquakes dataset is only a little worse than SAHP
and THP. Note that {\em our model only needs to fit tens of
parameters during training, compared to the hundreds of thousands
parameters needed for the neural networks}.

It is also notable that the 2-hidden is better than 1-hidden in terms of likelihood and accuracy while,
for RMSE, 2-hidden is worse than 1-hidden consistently. It indicates that separate hidden TPPs for each
type does increase the power to fit the evidence.
The RMSE of the 1-hidden model
is not maintained when increasing model capacity to the 2-hidden model, because we are fitting to
likelihood which is related but different than RMSE.
Note that on the
homicides data, our RMSE error for both 2-hidden and 1-hidden are
significantly better than those of prior work.

We also ran the real-world experiments for both a fully connected model and
a model with more than 2 hidden layers. The likelihood, RMSE, and accuracy
were not improved, because 
the single top layer is sufficient to correlate events across the
different types.
The time complexity analysis can be found in Appx.~\ref{sec:training-time}.

\section{RELATED WORK}
\paragraph{Neural Networks} Recent work uses neural networks to
directly model the CIF for a TPP, \textit{e.g.}, recurrent neural networks
\citep{du2016recurrent,mei2017neural} and
self-attention mechanism \citep{zuo2020transformer,zhang2020self}.
\citet{omi2019fully} model the cumulative hazard function. And, some others
attempt to model the conditional distribution of the inter-event times
\citep{ mehrasa2019variational,DBLP:conf/iclr/ShchurBG20}. All of these
methods assume the latent space is deterministically identified by a neural
network, which lacks the natural flexibility induced by the randomness of a
stochastic process. Most critically, a large number of parameters is
generally required.

\paragraph{Gaussian Processes (GP) Modulated Point Processes} Others
have
constructed GP-modulated point processes with the assumption that the
intensity functions are smooth
\citep[\textit{e.g.}][]{moller1998log,kottas2006dirichlet,kottas2007bayesian,adams2009tractable,rao2011gaussian,structcoxpp14_UAI,samo2015scalable,lloyd2015variational,donner2018efficient,aglietti2019structured}.
However, the intensity functions can experience sudden changes upon events
arrivals, \textit{e.g.}, a big earthquake can dramatically increase the
probability to have a small earthquake in the near future. And the integral
of the intensity function is not available in a closed form, requiring
approximations in its computation.  Others have imposed GP priors on the
CIFs of Hawkes processes
\citep{zhou2019efficient,zhang2019efficient,zhang2020variational,zhou2020efficient},
which assume the sudden changes can only happen at the times of the
observed events. All of these GP-modulated models are not able to be
stacked to construct a deep model built solely with point processes.  

\paragraph{Piecewise-constant Intensity Model (PCIM)} Another line of work it to use a PCIM
\citep{gunawardana2011model} or other variants \citep{weiss2013forest,lian2015multitask}. The history
of events is embedded into a piecewise-constant function, which is the intensity function itself or
some other function that can output the intensity. The size of time windows for the calculation of the
piecewise-constant function needs to be pre-determined.

\section{CONCLUSION}
We build a deep Neyman-Scott process (DN-SP) and use it to model real-world
event sequences. Different from the existing methods for Cox processes, we
are able to stack point processes in a hierarchical manner and do not
assume the intensity function is smooth in the space. We propose and test
an efficient MCMC posterior sampling algorithm for DN-SPs. Virtual events
as auxiliary variables help accelerate the mixing of our Markov chains.
With the fast posterior sampling, we are able to do inference for large
datasets. Our encouraging experiments results suggest that it is promising
to build a deep model with point processes. Future research directions
include constructing more complicated kernels, designing variational
inference algorithm, and generalizing the DN-SPs into more general spaces.

\subsubsection*{Acknowledgments}
Computational resources were provided through Office of Naval Research grant N00014-18-1-2252.

Data for this study came from
the Berkeley Digital Seismic Network (BDSN), doi:10/7932/BDSN;
the High Resolution Seismic Network (HRSN), doi:10.7932/HRSN; and
the Bay Area Regional Deformation Network (BARD), doi:10.7932/BARD,
all operated by the UC Berkeley Seismological Laboratory and
archived at the Northern California Earthquake Data center (NCEDC), doi: 10/7932/NCEDC.

\bibliography{sample_paper}


\clearpage
\appendix

\thispagestyle{empty}

\onecolumn \makesupplementtitle

\section{POSTERIOR PCIF} \label{sec:appen-posterior-pdf}
\begin{lemma}[\citet{kallenberg1984informal}] \label{lemma}
The PCIF for a point process \(\Xi\) defined on \(S\) with density \(f\) is 
\begin{equation*}
    \lambda_{\mathcal{P}}(t)=\frac{f(\xi\cup \{t\})}{f(\xi)},\ t \in S\backslash \xi,
\end{equation*}
with \(\xi\) as a realization of \(\Xi\) and \(a/0=0\) for \(a \geq 0\).
\end{lemma}
\setcounter{proposition}{0}
\begin{proposition}
The PCIF for the posterior point process of \(Z_{\ell,k}\) is
\begin{align}
    &\lambda_{\mathcal{P};\ell,k}(t)
    =\lambda_{\ell,k}(t) \prod_{i=1}^{K_{\ell-1}}\Bigg(\exp{\left(-\Phi_{(\ell,k) \rightarrow (\ell-1,i)}(T-t)\right)}
	\prod_{t_{\ell-1,i,j}>t}\frac{\lambda'_{\ell-1,i}(t_{\ell-1,i,j},t)}{\lambda_{\ell-1,i}(t_{\ell-1,i,j})}\Bigg), 
\end{align}
where \(\lambda'_{\ell-1,i}(x,y)=\lambda_{\ell-1,i}(x)+\phi_{(\ell,k) \rightarrow (\ell-1,i)}(x-y)\) and \(\Phi_{(\ell,k) \rightarrow (\ell-1,i)}(x)=\int_0^{x}\phi_{\bm{\theta}_{(\ell,k) \rightarrow (\ell-1,i)}}(\tau)d\tau\).
\end{proposition}
\begin{proof}
The \textit{p.d.f} to have an event at time \(t\) for \(Z_{\ell,k}\) is 
\begin{align*}
    p(t,\mathbf{x},\mathbf{z}_{1},\dots,\mathbf{z}_{\ell-1},\mathbf{z}_{\ell}/\{t\},\mathbf{z}_{\ell+1},\dots,\mathbf{z}_{L})
    = p(\mathbf{z}_{L})p(\mathbf{z}_{L-1}\mid\mathbf{z}_{L})\cdots p(\mathbf{x}\mid\mathbf{z}_1),
\end{align*}
where \(\mathbf{z}_{\ell}/\{t\}\) means there is no event at time \(t\) for \(Z_{\ell,k}\).

If there is no event at time \(t\), the \textit{p.d.f} conditional on the information of the point processes except time \(t\) is 
\begin{align*}
    p(\mathbf{x},\mathbf{z}_{1},\dots,\mathbf{z}_{\ell-1},\mathbf{z}_{\ell}/\{t\},\mathbf{z}_{\ell+1},\dots,\mathbf{z}_{L})
    =p(\mathbf{z}_{L})p(\mathbf{z}_{L-1}\mid\mathbf{z}_{L})\cdots p(\mathbf{z}_{\ell}/\{t\}\mid\mathbf{z}_{\ell+1})\cdots
	p(\mathbf{x}\mid\mathbf{z}_1).
\end{align*}

Thus, according to Lemma \ref{lemma}, the posterior PCIF to have an event at time \(t\) is
\begin{align*}
	\lambda_{\mathcal{P};\ell,k}(t) &=\frac{p(t,\mathbf{x},\mathbf{z}_{1},\dots,\mathbf{z}_{\ell-1},\mathbf{z}_{\ell}/\{t\},\mathbf{z}_{\ell+1},\dots,\mathbf{z}_{L})}{p(\mathbf{x},\mathbf{z}_{1},\dots,\mathbf{z}_{\ell-1},\mathbf{z}_{\ell}/\{t\},\mathbf{z}_{\ell+1},\dots,\mathbf{z}_{L})}\\
	&=\frac{p(\mathbf{z}_{\ell}/\{t\}\cup\{t\}\mid\mathbf{z}_{\ell+1})}{p(\mathbf{z}_{\ell}/\{t\}\mid\mathbf{z}_{\ell+1})}\\
	&=\lambda_{\ell,k}(t) \prod_{i=1}^{K_{\ell-1}}\Bigg(\exp{\left(-\Phi_{(\ell,k) \rightarrow
	(\ell-1,i)}(T-t)\right)}  \prod_{t_{\ell-1,i,j}>t}\frac{\lambda'_{\ell-1,i}(t_{\ell-1,i,j},t)}{\lambda_{\ell-1,i}(t_{\ell-1,i,j})}\Bigg).
\end{align*}
\end{proof}
\section{SAMPLER MOVES} \label{sec:sampler-moves}

Recall
\begin{align*}
\mathcal{P}=&\frac{p(\mathbf{x}, \mathbf{z}', \tilde{\mathbf{z}}')}{p(\mathbf{x}, \mathbf{z},
	\tilde{\mathbf{z}})}=\frac{p(\mathbf{z}'_{\ell,k}\mid\mathbf{z}_{\ell+1})\cdot
	\tilde{p}(\tilde{\mathbf{z}}'_{\ell,k}\mid\mathbf{z}_{\ell-1})}{p(\mathbf{z}_{\ell,k}\mid\mathbf{z}_{\ell+1})\cdot
	\tilde{p}(\tilde{\mathbf{z}}_{\ell,k}\mid\mathbf{z}_{\ell-1})}\cdot\frac{ p(\mathbf{z}_{\ell-1}\mid\mathbf{z}'_{\ell})\cdot
	\tilde{p}(\tilde{\mathbf{z}}_{\ell+1}\mid\mathbf{z}'_{\ell})}{ p(\mathbf{z}_{\ell-1}\mid\mathbf{z}_{\ell})\cdot
	\tilde{p}(\tilde{\mathbf{z}}_{\ell+1}\mid\mathbf{z}_{\ell})},
\end{align*}
where
\begin{align*}
  p(\mathbf{z}_{\ell-1}\mid\mathbf{z}_{\ell})&=\prod_{k=1}^{K_{\ell-1}}p(\mathbf{z}_{\ell-1,k}\mid\mathbf{z}_{\ell}),\\
  \tilde{p}(\tilde{\mathbf{z}}_{\ell+1}\mid\mathbf{z}_{\ell})&=\prod_{k=1}^{K_{\ell+1}}\tilde{p}(\tilde{\mathbf{z}}_{\ell+1,k}\mid\mathbf{z}_{\ell}).
\end{align*}

\subsection{Re-sample Virtual Events} \label{sec:appen-accep-resample-virtual}
Suppose we want to re-sample the virtual events for \(\tilde{Z}_{\ell,k}\), the proposal probability  is
\begin{equation*}
    q(\tilde{\mathbf{z}}_{\ell,k} \mid\mathbf{z}_{\ell-1})=\tilde{p}(\tilde{\mathbf{z}}_{\ell,k} \mid\mathbf{z}_{\ell-1}),
\end{equation*}
and the integrated detailed balance equation is
\begin{align*}
  &\int_A\int_{A'}p(\mathbf{x},\mathbf{z},\tilde{\mathbf{z}})q(\tilde{\mathbf{z}}_{\ell,k}' \mid\mathbf{z}_{\ell-1})\alpha(\tilde{\mathbf{z}}_{\ell,k},\tilde{\mathbf{z}}'_{\ell,k})d\tilde{\mathbf{z}}'_{\ell,k}d\tilde{\mathbf{z}}_{\ell,k}\\
  =&\int_A\int_{A'}p(\mathbf{x},\mathbf{z}',\tilde{\mathbf{z}}')q(\tilde{\mathbf{z}}_{\ell,k} \mid\mathbf{z}_{\ell-1})\alpha(\tilde{\mathbf{z}}'_{\ell,k},\tilde{\mathbf{z}}_{\ell,k})d\tilde{\mathbf{z}}'_{\ell,k}d\tilde{\mathbf{z}}_{\ell,k}, 
\end{align*}
where \(\tilde{\mathbf{z}}_{\ell,k}\in A\) and \(\tilde{\mathbf{z}}'_{\ell,k} \in A'\).

The proposal probability ratio is
\begin{align*}
    \mathcal{Q}=\frac{q(\tilde{\mathbf{z}}_{\ell,k} \mid\mathbf{z}_{\ell-1})}{q(\tilde{\mathbf{z}}'_{\ell,k} \mid\mathbf{z}_{\ell-1})}
	=\frac{\tilde{p}(\tilde{\mathbf{z}}_{\ell,k} \mid\mathbf{z}_{\ell-1})}{\tilde{p}(\tilde{\mathbf{z}}'_{\ell,k} \mid\mathbf{z}_{\ell-1})}.
\end{align*}
The likelihood ratio is
\begin{equation*}
    \mathcal{P}=\frac{p(\mathbf{x},\mathbf{z}',\tilde{\mathbf{z}}')}{p(\mathbf{x},\mathbf{z},\tilde{\mathbf{z}})}=\frac{\tilde{p}(\tilde{\mathbf{z}}'_{\ell,k}
	\mid\mathbf{z}_{\ell-1})}{\tilde{p}(\tilde{\mathbf{z}}_{\ell,k} \mid\mathbf{z}_{\ell-1})}.
\end{equation*}
The absolute value of the determinant of the Jacobian is 
\begin{equation*}
    \mathcal{J}=\abs{\frac{\partial (\tilde{\mathbf{z}}_{\ell,k}',\tilde{\mathbf{z}}_{\ell,k})}{\partial (\tilde{\mathbf{z}}_{\ell,k}', \tilde{\mathbf{z}}_{\ell,k})}}=1.
\end{equation*}
So the acceptance probability is 
\begin{equation*}
    \alpha(\tilde{\mathbf{z}}_{\ell,k}, \tilde{\mathbf{z}}'_{\ell,k})=\mathcal{P}\cdot\mathcal{Q}\cdot \mathcal{J}=1.
\end{equation*}
\subsection{Flip a Virtual Event to a Real Event} \label{sec:appen-flip-virtual-to-real}
Suppose we want to flip a virtual event \(\tilde{t}_{\ell,k,j}\) to a real event \(t_{\ell,k,m_{\ell,k}+1}\). 

The number of events for the current state and the proposed state is \(m_{\ell,k}+\tilde{m}_{\ell,k}\). The
probabilities to pick \(\tilde{t}_{\ell,k,j}\) and  \(t_{\ell,k,m_{\ell,k}+1}\) are both
\(1/(m_{\ell,k}+\tilde{m}_{\ell,k})\), as the total number of events does not change.  Hence, the proposal probability ratio is \[\mathcal{Q}=\frac{1/(m_{\ell,k}+\tilde{m}_{\ell,k})}{1/(m_{\ell,k}+\tilde{m}_{\ell,k})}=1.\]
The likelihood ratio is 
\begin{align*}
    \mathcal{P}_v=&\frac{\lambda_{\ell,k}(\tilde{t}_{\ell,k,j})}{\tilde{\lambda}_{\ell,k}(\tilde{t}_{\ell,k,j})}\cdot\frac{
	    p(\mathbf{z}_{\ell-1}\mid\mathbf{z}'_{\ell})\cdot \tilde{p}(\tilde{\mathbf{z}}_{\ell+1}\mid\mathbf{z}'_{\ell})}{
	    p(\mathbf{z}_{\ell-1}\mid\mathbf{z}_{\ell})\cdot \tilde{p}(\tilde{\mathbf{z}}_{\ell+1}\mid\mathbf{z}_{\ell})}\\
    =&\frac{\lambda_{\ell,k}(\tilde{t}_{\ell,k,j})}{\tilde{\lambda}_{\ell,k}(\tilde{t}_{\ell,k,j})}\\
    &\cdot\prod_{i=1}^{K_{\ell-1}}\Bigg(\exp\left(-\Phi_{\bm{\theta}_{(\ell,k)\rightarrow(\ell-1,i)}}(T-\tilde{t}_{\ell,k,j})\right)
	\frac{\prod_{t_{\ell-1,i,j}>\tilde{t}_{\ell,k,j}}\lambda'_{\ell-1,i}(t_{\ell-1,i,j},\tilde{t}_{\ell,k,j})}{\prod_{t_{\ell-1,i,j}>\tilde{t}_{\ell,k,j}}\lambda_{\ell-1,i}(t_{\ell-1,i,j})}\Bigg)\\
    &\cdot
	\prod_{i=1}^{K_{\ell+1}}\Bigg(\exp\left(-\tilde{\Phi}_{\tilde{\bm{\theta}}_{(\ell,k)\rightarrow(\ell+1,i)}}(\tilde{t}_{\ell,k,j})\right)
	\frac{\prod_{\tilde{t}_{\ell+1,i,j}<\tilde{t}_{\ell,k,j}}\tilde{\lambda}'_{\ell+1,i}(\tilde{t}_{\ell+1,i,j},\tilde{t}_{\ell,k,j})}{\prod_{\tilde{t}_{\ell+1,i,j}<\tilde{t}_{\ell,k,j}}\tilde{\lambda}_{\ell+1,i}(\tilde{t}_{\ell+1,i,j})}\Bigg) ,
\end{align*}
where
\(\tilde{\Phi}_{\tilde{\bm{\theta}}_{(\ell,k)\rightarrow(\ell+1,i)}}(x)=\int_0^x\tilde{\phi}_{\tilde{\bm{\theta}}_{(\ell,k)\rightarrow(\ell+1,i)}}(\tau)d\tau\),
\(\lambda'_{\ell-1,i}(x,y)=\lambda_{\ell-1,i}(x)+\phi_{\bm{\theta}_{(\ell,k) \rightarrow (\ell-1,i)}}(x-y)\), and
\(\tilde{\lambda}'_{\ell+1,i}(x,y)=\tilde{\lambda}_{\ell+1,i}(x)+\tilde{\phi}_{\tilde{\bm{\theta}}_{(\ell,k)
\rightarrow (\ell+1,i)}}(y-x)\).

So the acceptance probability is 
\[\alpha=\min(1,\mathcal{P}_v).\]
\subsection{Flip a Real Event to a Virtual Event} \label{sec:appen-flip-real-to-virtual}
Suppose we want to flip a real event \(t_{\ell,k,j}\) to a virtual event \(\tilde{t}_{\ell,k,\tilde{m}_{\ell,k}+1}\). 

Similar to Section \ref{sec:appen-flip-virtual-to-real}, the proposal probability ratio is \(\mathcal{Q}=1\).

The likelihood ratio is 
\begin{align*}
    \mathcal{P}_r=&\frac{\tilde{\lambda}_{\ell,k}(t_{\ell,k,j})}{\lambda_{\ell,k}(t_{\ell,k,j})}\cdot \frac{
	    p(\mathbf{z}_{\ell-1}\mid\mathbf{z}'_{\ell})\cdot \tilde{p}(\tilde{\mathbf{z}}_{\ell+1}\mid\mathbf{z}'_{\ell})}{
	    p(\mathbf{z}_{\ell-1}\mid\mathbf{z}_{\ell})\cdot \tilde{p}(\tilde{\mathbf{z}}_{\ell+1}\mid\mathbf{z}_{\ell})}\\
    =&\frac{\tilde{\lambda}_{\ell,k}(t_{\ell,k,j})}{\lambda_{\ell,k}(t_{\ell,k,j})}\\
    &\cdot\prod_{i=1}^{K_{\ell-1}}\Bigg(\exp\left(\Phi_{\bm{\theta}_{(\ell,k)\rightarrow(\ell-1,i)}}(T-t_{\ell,k,j})\right)
	\frac{\prod_{t_{\ell-1,i,j}>t_{\ell,k,j}}\lambda'_{\ell-1,i}(t_{\ell-1,i,j},t_{\ell,k,j})}{\prod_{t_{\ell-1,i,j}>t_{\ell,k,j}}\lambda_{\ell-1,i}(t_{\ell-1,i,j})}\Bigg)\\
    &\cdot
	\prod_{i=1}^{K_{\ell+1}}\Bigg(\exp\left(\tilde{\Phi}_{\tilde{\bm{\theta}}_{(\ell,k)\rightarrow(\ell+1,i)}}(t_{\ell,k,j})\right)
	\frac{\prod_{\tilde{t}_{\ell+1,i,j}<t_{\ell,k,j}}\tilde{\lambda}'_{\ell+1,i}(\tilde{t}_{\ell+1,i,j},t_{\ell,k,j})}{\prod_{\tilde{t}_{\ell+1,i,j}<t_{\ell,k,j}}\tilde{\lambda}_{\ell+1,i}(\tilde{t}_{\ell+1,i,j})}\Bigg) ,
\end{align*}
where \(\lambda'_{\ell-1,i}(x,y)=\lambda_{\ell-1,i}(x)-\phi_{\bm{\theta}_{(\ell,k) \rightarrow (\ell-1,i)}}(x-y)\), and
\(\tilde{\lambda}'_{\ell+1,i}(x,y)=\tilde{\lambda}_{\ell+1,i}(x)-\tilde{\phi}_{\tilde{\bm{\theta}}_{(\ell,k) \rightarrow (\ell+1,i)}}(y-x)\).

So the acceptance probability is 
\[\alpha=\min(1,\mathcal{P}_r).\]
\subsection{Swap a Real Event and a Virtual Event}
Suppose we want to flip a virtual event \(\tilde{t}_{\ell,k,j}\) to a real event \(t_{\ell,k,m_{\ell,i}+1}\) and we also want to flip a real event \(t_{\ell,k,j}\) to a virtual event \(\tilde{t}_{\ell,k,\tilde{m}_{\ell,k}+1}\).

The probability to pick \(\tilde{t}_{\ell,k,j}\) and \(t_{\ell,k,j}\) is \(1/m_{\ell,k}\cdot 1/\tilde{m}_{\ell,k}\), the same for the probability to pick the real and virtual events in the reverse move. Thus, the proposal probability ratio is \(\mathcal{Q}=1\).

The likelihood ratio is
\begin{equation*}
    \mathcal{P}=\mathcal{P}_v\cdot \mathcal{P}_r\,\,.
\end{equation*}
So the acceptance probability is 
\[\alpha=\min(1,\mathcal{P}).\]
\section{OPTIMIZATION} \label{sec:appen-grad}
\subsection{Derivatives \textit{w.r.t} the Parameters for VPPs} \label{sec:appen-opt-vpps}
For simplicity, we omit the data index \(n\). The likelihood of the parameters for the VPPs \textit{w.r.t} the real events is
\begin{align*}
\tilde{\llh}=\sum_{\ell=1}^L\sum_{k=1}^{K_\ell}\Bigg(\sum_{j=1}^{\tilde{m}_{\ell,k}} \log \tilde{\lambda}_{\ell,k}(t_{\ell,k,j})-\int_0^T\tilde{\lambda}_{\ell,k}(t)dt\Bigg),
\end{align*}
where 
\begin{align*}
    \int_0^T\tilde{\lambda}_{\ell,k}(t)dt=\tilde{\mu}_{\ell,k}T+\sum_{i=1}^{K_{\ell-1}}\sum_{j=1}^{m_{\ell-1,i}}\tilde{\Phi}_{\tilde{\bm{\theta}}_{(\ell-1,i)\rightarrow(\ell,k)}}(t_{\ell-1,i,j}).
\end{align*}
The integral of the virtual kernel function is
\begin{align*}
    &\tilde{\Phi}_{\tilde{\bm{\theta}}_{(\ell-1,i)\rightarrow(\ell,k)}}(t)=\int_0^t\tilde{\phi}_{\tilde{\bm{\theta}}_{(\ell,i)\rightarrow(\ell+1,k)}}(\tau)d\tau=\int_0^t \tilde{p} \frac{\tilde{\beta}^{\tilde{\alpha}}}{\Gamma(\tilde{\alpha})}\tau^{\tilde{\alpha}-1}e^{-\tilde{\beta} \tau}d\tau=\tilde{p} \frac{1}{\Gamma(\tilde{\alpha})} \gamma(\tilde{\alpha}, \tilde{\beta} t)
\end{align*}
with \(\tilde{p}=\tilde{p}_{(\ell-1,i)\rightarrow(\ell,k)}\), \(\tilde{\alpha}=\tilde{\alpha}_{(\ell-1,i)\rightarrow(\ell,k)}\), \(\tilde{\beta}=\tilde{\beta}_{(\ell-1,i)\rightarrow(\ell,k)}\).

The partial derivative of \(\tilde{\llh}\) \textit{w.r.t} \(\tilde{\mu}_{\ell,k}\) is
\begin{equation*}
    \partial_{\tilde{\mu}_{\ell,k}}\tilde{\llh}=\sum_{j=1}^{\tilde{m}_{\ell,k}}\frac{1}{\tilde{\lambda}_{\ell,k}(t_{\ell,k,j})}-T.
\end{equation*}
In the following section, we use \(\tilde{\bm{\theta}}\) to denote \(\tilde{\bm{\theta}}_{(\ell-1,*)\rightarrow(\ell,k)}\).

The partial derivative of \(\tilde{\llh}\) \textit{w.r.t} \(\tilde{p}\) is
\begin{align*}
    \partial_{\tilde{p}}\tilde{\llh}=&\partial_{\tilde{p}}\Bigg(\sum_{j=1}^{\tilde{m}_{\ell,k}} \log \tilde{\lambda}_{\ell,k}(t_{\ell,k,j})-\sum_{i=1}^{K_{\ell-1}}\sum_{j=1}^{m_{\ell-1,i}}\tilde{\Phi}_{\tilde{\bm{\theta}}}(t_{\ell-1,i,j})\Bigg)\\
    =&\sum_{j=1}^{\tilde{m}_{\ell,k}} \frac{\partial_{\tilde{p}}\phi_{\tilde{\bm{\theta}}_{\tilde{\bm{\theta}}}}(t_{\ell,k,j})}{\tilde{\lambda}_{\ell,k}(t_{\ell,k,j})}-\sum_{i=1}^{K_{\ell-1}}\sum_{j=1}^{m_{\ell-1,i}}\partial_{\tilde{p}}\tilde{\Phi}_{\tilde{\bm{\theta}}_{\tilde{\bm{\theta}}}}(t_{\ell-1,i,j}),
\end{align*}
where
\begin{align*}
    \partial_{\tilde{p}}\phi_{\tilde{\bm{\theta}}}(t)=\frac{\tilde{\beta}^{\tilde{\alpha}}}{\Gamma(\tilde{\alpha})}t^{\tilde{\alpha}-1}e^{-\tilde{\beta} t}\text{, } \partial_{\tilde{p}}\tilde{\Phi}_{\tilde{\bm{\theta}}}(t)=\frac{1}{\Gamma(\tilde{\alpha})} \gamma(\tilde{\alpha}, \tilde{\beta} t).
\end{align*}
The partial derivative of \(\tilde{\llh}\) \textit{w.r.t} \(\tilde{\alpha}\) is
\begin{align*}
    &\partial_{\tilde{\alpha}}\tilde{\llh}=\sum_{j=1}^{\tilde{m}_{\ell,k}} \frac{\partial_{\tilde{\alpha}}\phi_{\tilde{\bm{\theta}}}(t_{\ell,k,j})}{\tilde{\lambda}_{\ell,k}(t_{\ell,k,j})}-\sum_{i=1}^{K_{\ell-1}}\sum_{j=1}^{m_{\ell-1,i}}\partial_{\tilde{\alpha}}\tilde{\Phi}_{\tilde{\bm{\theta}}}(t_{\ell-1,i,j}),
\end{align*}
where
\begin{align*}
    \partial_{\tilde{\alpha}}\phi_{\tilde{\bm{\theta}}}(t)=&\tilde{p} \frac{(\tilde{\beta} t) ^{\tilde{\alpha}-1}\ln (\tilde{\beta} t) \Gamma(\tilde{\alpha})-(\tilde{\beta} t)^{\tilde{\alpha}-1}\Psi(\tilde{\alpha})\Gamma(\tilde{\alpha})}{\Gamma^2(\tilde{\alpha})}\tilde{\beta} e^{-\tilde{\beta} t}=\tilde{p} (\tilde{\beta} t)^{\tilde{\alpha}-1}\frac{\ln(\tilde{\beta} t)-\Psi(\tilde{\alpha})}{\Gamma(\tilde{\alpha})}\tilde{\beta} e^{-\tilde{\beta} t},\\
    \partial_{\tilde{\alpha}} \Phi_{\tilde{\bm{\theta}}}(t)=&\tilde{p}\left(-\frac{\Psi(\tilde{\alpha})}{\Gamma(\tilde{\alpha})}\gamma(\tilde{\alpha}, \tilde{\beta} t)+\frac{1}{\Gamma(\tilde{\alpha})}\frac{\partial \gamma(\tilde{\alpha}, \tilde{\beta} t)}{\partial \tilde{\alpha}}\right)\\
=&\tilde{p} \left(-\frac{\Psi(\tilde{\alpha})}{\Gamma(\tilde{\alpha})}\gamma(\tilde{\alpha}, \tilde{\beta} t)+\frac{1}{\Gamma(\tilde{\alpha})}\frac{\partial (\Gamma(\tilde{\alpha})-\Gamma(\tilde{\alpha},\tilde{\beta} t))}{\partial \tilde{\alpha}}\right)\\
=&\tilde{p} \Bigg(-\frac{\Psi(\tilde{\alpha})}{\Gamma(\tilde{\alpha})}\gamma(\tilde{\alpha}, \tilde{\beta} t)+\frac{1}{\Gamma(\tilde{\alpha})} \cdot\left(\Psi(\tilde{\alpha}) \Gamma(\tilde{\alpha}) - \ln (\tilde{\beta} t)\Gamma(\tilde{\alpha}, \tilde{\beta} t)-\tilde{\beta} t \cdot T(3, \tilde{\alpha}, \tilde{\beta} t)\right)\Bigg),\\
\end{align*}
and \(T(m,s,x)\) is a special case of the Meijer G-function \[T(m,s,x)=G^{m,0}_{m-1,m}\left(\begin{array}{c}
     \underbrace{0,0,\cdots,0}_{m-1}  \\
     \underbrace{s-1,-1,\cdots,-1}_{m} 
\end{array}\middle \vert x\right).\]
However, it is expensive and numerically unstable to directly calculate Meijer G-function, so we use the first order finite difference to approximate the derivative.

The partial derivative of \(\tilde{\llh}\) \textit{w.r.t} \(\tilde{\beta}\) is
\begin{align*}
	\partial_{\tilde{\beta}}\tilde{\llh}&=\sum_{j=1}^{\tilde{m}_{\ell,k}}
	\frac{\partial_{\tilde{\beta}}\phi_{\tilde{\bm{\theta}}}(t_{\ell,k,j})}{\tilde{\lambda}_{\ell,k}(t_{\ell,k,j})}-\sum_{i=1}^{K_{\ell-1}}\sum_{j=1}^{m_{\ell-1,i}}\partial_{\tilde{\beta}}\tilde{\Phi}_{\tilde{\bm{\theta}}}(t_{\ell-1,i,j}),
	\\
	\intertext{where}
	\partial_{\tilde{\beta}} \phi_{\tilde{\bm{\theta}}}(t)&=\tilde{p} /
	\Gamma(\tilde{\alpha})t^{\tilde{\alpha}-1}(\tilde{\alpha} \tilde{\beta}^{\tilde{\alpha}-1}e^{-\tilde{\beta} t} -
	\tilde{\beta}^{\tilde{\alpha}}e^{-\tilde{\beta} t}t)=\tilde{p}
	/\Gamma(\tilde{\alpha})t^{\tilde{\alpha}-1}\tilde{\beta}^{\tilde{\alpha}-1}e^{-\tilde{\beta}
	t}(\tilde{\alpha}-\tilde{\beta} t)\,\,, \\
	\partial_{\tilde{\beta}} \Phi_{\tilde{\bm{\theta}}}(t)&=\tilde{p}\frac{1}{\Gamma(\tilde{\alpha})}\frac{\partial
	\gamma(\tilde{\alpha}, \tilde{\beta} t)}{\partial \tilde{\beta}}=\tilde{p}
	\frac{1}{\Gamma(\tilde{\alpha})}(\tilde{\beta} t)^{\tilde{\alpha}-1}e^{-\tilde{\beta} t}\cdot t \,\,.
\end{align*}
We use softplus function to make sure the parameters are all positive.
\subsection{Maximization and Derivatives \textit{w.r.t} the Parameters for RPPs} \label{sec:appen-opt-rpps}
The likelihood of the parameters for \(\mathbf{Z}_{n}\) \textit{w.r.t} the real events is
\begin{align*}
	\llh&=\sum_{\ell=0}^L\sum_{k=1}^{K_\ell}\Bigg(\sum_{j=1}^{m_{n,\ell,k}} \log \lambda_{n,\ell,k}(t_{n,\ell,k,j})-\int_0^T\lambda_{n,\ell,k}(t)dt\Bigg),
\intertext{where}
\int_0^T\lambda_{n,L,k}(t)dt&=\mu_{n,k}T\,\,,\\
    \int_0^T\lambda_{n,\ell,k}(t)dt&=\sum_{i=1}^{K_{\ell+1}}\sum_{j=1}^{m_{n,\ell+1,i}}\Phi_{\bm{\theta}_{(\ell+1,i)\rightarrow(\ell,k)}}(t_{n,\ell+1,i,j})\text{ for }0\leq \ell \leq L-1\,\,.
\end{align*}
The maximizing value for \(\mu_{n,L,k}\) is 
\begin{equation*}
    \mu_{n,L,k}=\frac{m_{n,L,k}}{T}.
\end{equation*}
The functional forms for the derivatives of the other parameters of the kernel functions are the same  as the forms for VPPs.
\section{FULL BAYESIAN INFERENCE} \label{sec:appen-full-bayesian}
For simplicity, we use \(\bm{\theta}\) to denote all of the hyperparameters. We need sample \(\bm{\theta}\) from the posterior distribution
\[p(\bm{\theta}\mid\mathbf{x},\mathbf{z}) = \frac{p(\mathbf{x},\mathbf{z},\bm{\theta})}{p(\mathbf{x},\mathbf{z})}.\]

Each time we uniformly select a hyperparameter \(\theta\) from \(\bm{\theta}\) and the acceptance ratio is
\begin{align*}
	\mathcal{A}&=\min(\alpha,1),\\
	\intertext{where}
	\alpha&=\frac{p(\theta'\mid\mathbf{x},\mathbf{z})q(\theta'\rightarrow
	\theta)}{p(\theta\mid\mathbf{x},\mathbf{z})q(\theta\rightarrow
	\theta')}=\frac{p(\theta')p(\mathbf{x},\mathbf{z}\mid\theta')q(\theta'\rightarrow \theta)}{p(\theta)p(\mathbf{x},\mathbf{z}\mid\theta)q(\theta\rightarrow \theta')},
\end{align*}
\(\theta'\) is the proposal, and \(\theta\) is the current value.

We assume the prior distributions for each parameter is Gamma distribution,  and use \(h\) to denote the shape, \(c\) to denote the scale, then
\begin{align*}
	p(\theta) &= \text{Gamma}(\theta ; h, c), \\
	q(\theta'\rightarrow \theta)&=\text{Gamma}(\theta;h, \theta'/h), \\
	q(\theta \rightarrow \theta')&=\text{Gamma}(\theta';h, \theta/h),
\end{align*}
and
\begin{align*}
    \alpha=&\frac{\cancel{1/(\Gamma(h)c^h)(\theta')^{h-1}}\exp(-\theta'/c)\cdot
	1/\left(\cancel{\Gamma(h)}(\theta'/h)^h\right)\cancel{\theta^{h-1}}\exp(-\theta/(\theta'/h))\cdot
	p(\mathbf{x},\mathbf{z}\mid\theta')}{\cancel{1/(\Gamma(h)c^h)\theta^{h-1}}\exp(-\theta/c)\cdot
	1/\left(\cancel{\Gamma(h)}(\theta/h)^h\right)\cancel{(\theta')^{h-1}}\exp(-\theta'/(\theta/h))\cdot p(\mathbf{x},\mathbf{z}\mid\theta)}\\
    =&\frac{\theta^h \exp(-\theta'/c-\theta/(\theta'/h))\cdot
	p(\mathbf{x},\mathbf{z}\mid\theta')}{(\theta')^h\exp(-\theta/c-\theta'/(\theta/h))\cdot p(\mathbf{x},\mathbf{z}\mid\theta)}.
\end{align*}
\section{SPATIAL BIRTH-AND-DEATH ALGORITHM} \label{appen:sbd}
We attempt to construct a spatial birth-death process to simulate the posterior distribution of hidden TPPs. We need to
construct two proposals: birth and death. We use a birth to add an event and a death to delete an event.

Spatial birth-and-death \citep{preston1977spatial} is a continuous-time Markov process. The detailed balance equation is 
\begin{equation}
    p(\mathbf{z}_{\ell,k}\mid\mathbf{x})b(\mathbf{z}_{\ell,k},t)=p(\mathbf{z}_{\ell,k}\cup \{t\}\mid\mathbf{x})d(\mathbf{z}_{\ell,k}\cup \{t\},t) \label{eq:b-d-balance},
\end{equation}
where \(b(\mathbf{z}_{\ell,k},t)\) is the birth rate to add a new event at time \(t\) to the current hidden events set \(\mathbf{z}_{\ell,k}\) and \(d(\mathbf{z}_{\ell,k},t)\) is the death rate for removing an event with time \(t\).

A common way to determine the birth rate \(b(\mathbf{z}_{\ell,k},t)\) is to make it proportional to the PCIF as in
Eq.~\ref{eq:posterior-pcif} and the death rate to be a constant number
\citep{ripley1977modelling,baddeley1989nearest,moller1989rate}. However, it would be very hard to calculate the total
birth rate exactly, as it would require an integral of the product terms in Eq.~\ref{eq:posterior-pcif}. If,
instead, we try to find an upper bound for the PCIF and then use thinning to get the samples for birth process, it
would have far too many attempted jumps rejected for the birth stage.

Similar to \citet{van2001extrapolating}, we can make the birth rate to be
\begin{equation}
    b(\mathbf{z}_{\ell,k},t)= \tilde{\mu}_{\ell,k}+\sum_{i=1}^{K_{\ell-1}}\sum_{t_{\ell-1,i,j}>t}\tilde{\phi}_{\tilde{\bm{\theta}}_{(\ell-1,i) \rightarrow (\ell,k)}}(t_{\ell-1,i,j}-t). \label{eq:birth-rate}
\end{equation}
The only difference from \citet{van2001extrapolating} is we divide the birth rate by \(\lambda_{\ell,k}(\cdot)\) and the birth rate not only depends on the evidence, but also depends on the posterior samples.

To satisfy the detailed balance Eq.~\ref{eq:b-d-balance}, the death rate \(d(\mathbf{z}_{\ell,k}\cup t, t)\) needs to be
\begin{align}
    d(\mathbf{z}_{\ell,k}\cup t, t)=&\frac{b(\mathbf{z}_{\ell,k},t)}{\lambda_{\mathcal{P};\ell,k}(t)}\nonumber\\
    =&\frac{\tilde{\mu}_{\ell,k}+\sum_{i=1}^{K_{\ell-1}}\sum_{t_{\ell-1,i,j}>t}\tilde{\phi}_{\tilde{\bm{\theta}}_{(\ell-1,i) \rightarrow (\ell,k)}}(t_{\ell-1,i,j}-t)}{\lambda_{\ell,k}(t) \prod_{i=1}^{K_{\ell-1}}\Bigg(\exp{\left(-\Phi_{(\ell,k) \rightarrow (\ell-1,i)}(T-t)\right)} \prod_{t_{\ell-1,i,j}>t}\frac{\lambda'_{\ell-1,i}(t_{\ell-1,i,j},t)}{\lambda_{\ell-1,i}(t_{\ell-1,i,j})}\Bigg)}, \label{eq:death-rate}
\end{align}
where \(\lambda'_{\ell-1,i}(x,y)=\lambda_{\ell-1,i}(x)+\phi_{\theta_{(\ell,k)\rightarrow(\ell-1,i)}}(x-y)\).

The total birth rate is 
\begin{equation}
    \beta(\mathbf{z})=\sum_{\ell,k}\int_0^Tb(\mathbf{z}_{\ell,k},t)dt=\sum_{\ell,k}\left(\tilde{\mu}_{\ell,k}T + \sum_{i=1}^{K_{\ell-1}}\sum_{t_{\ell-1,i,j}}\tilde{\Phi}_{\tilde{\theta}_{(\ell-1,i)\rightarrow(\ell,k)}}(t_{\ell-1,i,j})\right), 
    \label{eq:total-birth-rate}
\end{equation}
and the total death rate is
\begin{equation*}
    \delta(\mathbf{z})=\sum_{\ell,k}\left(\sum_{j}d(\mathbf{z}_{\ell,k},t_{\ell,k,j})\right).
\end{equation*}

\textbf{Open problem:} \textit{Does the SB\&D with the birth rate as in Eq.~\ref{eq:birth-rate} and the death rate as in Eq.~\ref{eq:death-rate} converge to an invariant distribution?}

The SB\&D given in \citet{van2001extrapolating} is guaranteed to converge to an invariant distribution as it satisfies a sufficient condition that the total birth rate is bounded from above and the total death rate is bounded from below. However, the total birth rate in Eq.~\ref{eq:total-birth-rate} is finite but not bounded in our case, which violates the sufficient condition satisfied in \citet{van2001extrapolating}. Please refer to \citet{moller2003statistical} for more details.

Therefore, we cannot guarantee this method will converge, and instead use
the MCMC method in the main section of this paper.

\section{INFERENCE ALGORITHM} \label{sec:inf-alg}
\begin{algorithm}[H]
  \caption{MCEM for DN-SPs}
  \label{alg:mcem}
\begin{algorithmic}
  \STATE {\bfseries Input:} data $\{\mathbf{x}_n\}$, model \(\mathcal{M}\) 
  \STATE {\bfseries Initialization:} base rates \(\{\bm{\mu}_n\}\), kernel parameters $\bm{\theta}$, virtual base rates \(\{\tilde{\bm{\mu}}_n\}\), virtual kernel parameters $\bm{\tilde{\theta}}$, initial states for Markov chains.
  \REPEAT
  \STATE \(\mathbf{z}_n^{(1)},\dots,\mathbf{z}_n^{(\mathcal{S})} \sim p(\mathbf{Z}_n\mid\mathbf{x}_n;\bm{\mu}_n,\bm{\theta})\) by MCMC
  \STATE Maximize base rates based on Eq.~\ref{eq:update-real-mu}
  \STATE Use Adam to optimize Eqs.~\ref{eq:update-real-kernel}, \ref{eq:update-virtual-mu}, and \ref{eq:update-virtual-kernel}
  \STATE Record \(\mathbf{z}_n^{(\mathcal{S})}\) as the initial state for the Markov chain
  \UNTIL{the expected log-likelihood in Eq.~\ref{eq:ll-to-optimize} get converged}
\end{algorithmic}
\end{algorithm}

\section{EXPERIMENTS DETAILS} \label{sec:exp-details}
\subsection{Datasets} \label{sec:datasets-stats}
\paragraph{Retweets} \citep{zhao2015seismic} The retweets dataset collected sequences of tweets streams. Each sequence contains the times and types for some follow-up retweets. We use the same dataset as used in NHP. The retweets are grouped into three types (small, medium and large) according to the number of followers of the users who owned the retweets.

\paragraph{Earthquakes} \citep{ncedc,bdsn,hrsn,brad} We collected the times and magnitudes for earthquakes between 01/01/2014 00:00:00 and 01/01/2020 00:00:00 in the region spanning between \(34.5^{\circ}\) and \(43.2^{\circ}\) latitude and between \(-126.00^{\circ}\) and \(-117.76^{\circ}\) longitude. If the magnitude of a earthquake is smaller than 1, we classify it as a small earthquake, otherwise a large earthquake. 

\paragraph{Homicides} \citep{chicagocrime} This dataset contains the
times for homicides that occurred at five contiguous districts (007-011)
with the most homicides in Chicago from 01/01/2001 00:00:00 to 01/01/2020
00:00:00. The type for an event is the district where the homicide occurred. The terms of use can be found at \url{https://www.chicago.gov/city/en/narr/foia/data_disclaimer.html}.

The number of types, the number of total events and the number of sequences for each dataset are summarized in Table \ref{table:dataset-stats}.

\begin{table}[H]
\caption{Datasets Statistics}
\label{table:dataset-stats}
\begin{center}
\begin{small}
\begin{sc}
\begin{tabular}{lcrrrr}
\toprule
\multirow{2}{*}{Datasets} & \multirow{2}{*}{\# types} & \multirow{2}{*}{\# events} & \multicolumn{3}{c}{\# sequences}\\
\cmidrule{4-6}
&  & & train &validation &test \\
\midrule
Retweets &3    &2610102 &20000 & 2000 & 2000 \\
Earthquakes &2    &156743  & 209 & 53 & 53 \\
Homicides &5    &3956 & 6 & 2 & 2 \\
\bottomrule
\end{tabular}
\end{sc}
\end{small}
\end{center}
\vskip -0.1in
\end{table}
\subsection{Training and Testing}
The probabilities of the moves for \textit{Move 1}, \textit{Move 2}, and \textit{Move 3} are 0.2, 0.6, and 0.2 respectively.
We train the models using Algorithm \ref{alg:mcem} to get the parameters of the kernel and virtual kernel functions.

During testing, the parameters of the kernel functions and virtual kernel
functions are fixed. We update the base rates according to Eqs.~\ref{eq:update-real-mu} and \ref{eq:update-virtual-mu}. The posterior samples for RPPs are collected to calculate the expectation of the log-likelihood per event.
For each sequence in the evidence, there is an event at the end. To better capture the last event, we assume there is a synthetic real event for each hidden TPP at the end. This synthetic real event is only involved with the calculation of the intensity functions for the VPPs.  
\subsection{Likelihood}
For a neural network, the CIF has a parametric form \(\lambda_{f_{\Theta}(\mathbf{x})}(t)\) determined by a function \(f_{\Theta}\) and the data \(\mathbf{x}\). The parameters \(\Theta\) for \(f_{\Theta}\) are learned during training. For testing, the CIF \(\lambda_{f_{\Theta}(\mathbf{x})}(t)\) is determined by the testing data \(\mathbf{x}\) and \(f_{\Theta}\). The log-likelihood is \(\log P(\mathbf{x}\mid \lambda_{f_{\Theta}(\mathbf{x})}(t))\).

For a DN-SP, \(\Theta\), the parameters for the kernels, are learned during
training, and fully specify \(P(\mathbf{Z}\mid \mathbf{x})\). We use a random function \(f_{\Theta}\) to determine the CIF. The distribution of \(f_{\Theta}(\mathbf{x})\) is fully determined by \(P(\mathbf{Z}\mid \mathbf{x})\), and thus is fully determined by \(\Theta\) and \(\mathbf{x}\). We calculate the expected value of the log-likelihood \(\mathbb{E}_{f_{\Theta}(\mathbf{x})}[\log P(\mathbf{x}\mid \lambda_{f_{\Theta}(\mathbf{x})}(t))]=\mathbb{E}_{\mathbf{Z}\sim P(\mathbf{Z}\mid \mathbf{x})}[\log P(\mathbf{x}\mid \mathbf{Z})]\) to compare with the baselines.

We do not compare the marginal log-likelihood \(\log
\mathbb{E}_{\mathbf{Z}\sim P(\mathbf{Z})}[ P(\mathbf{x}\mid \mathbf{Z})]\)
of our model with the log-likelihood of the neural-network-based models
because there are no efficient methods for estimating this expectation,
as ``forward sampling'' with \(P(\mathbf{Z})\) is prohibitively inefficient and samples from
\(P(\mathbf{Z}\mid \mathbf{x})\), which our method is designed to generate,
cannot be used (as the importance sampling ratio weight cannot be
calculated).

Similarly, it is difficult to calculate the marginal likelihood for MTPP and the log-likelihood for MTPP is the evidence lower bound per event.


 To
address the difference in likelihood calculations across methods, we also
provide two additional metrics: predictive accuracy and root mean squared
error, which are handled in exactly the same fashion for all methods (see
Appx.~\ref{sec:pred-details}).

\subsection{Prediction}
\label{sec:pred-details}
For prediction, each method predicts the time and type of the next event,
conditioned on all events prior to it.  This process is repeated for
every event in the testing data (slowly increasing the conditioning set
for each new prediction).

Instead of calculating an infinite integral as in NHP and SAHP, or using an additional layer of a neural network as in THP, we simply do forward sampling to predict the time and type for the next future event, as we have an explicit formula for the intensity. 

We predict the time for the next future event from the beginning to the end. After we get to the convergence of our Markov chain, we draw the samples for the next future event \(e_i=(t_i,k_i)\) conditional on the current state of the Markov chain.
Suppose the samples for the next future event \(e_i\) conditional on the history \(\mathcal{H}_{i-1}=\{(t_1,k_1), (t_2,k_2), (t_3,k_3), \cdots, (t_{i-1},k_{i-1})\}\) are \((t^1_i, k^1_i), (t^2_i, k^2_i),\dots,(t^\mathcal{S}_i,k^\mathcal{S}_i)\),
where \(\{t^j_i\}_{j=1}^{\mathcal{S}}\) are the times of the samples for the future event \(e_i\) and \(\{k^j_i\}_{j=1}^{\mathcal{S}}\) are the types of the future event \(e_i\). The prediction for the future event time is \(\hat{t}_i=\frac{1}{\mathcal{S}}\sum_{j=1}^{\mathcal{S}}t^j_i\) and the type prediction is \(\hat{k}_i=\argmax_{k \in \{1,\dots,K_0\}}\sum_{j=1}^{\mathcal{S}}\mathds{1}_k(k^j_i)\), where  \(\mathds{1}_k(k_i^j)\) is the indicator function which is equal to 1 \textit{iff} \(k=k_i^j\). Then we calculate the root mean squared error (RMSE) for the time prediction and accuracy for the type prediction.

After the prediction for \(e_i\), we use the current state of the Markov chain for \(\mathcal{H}_{i-1}\) as the initial state for the Markov chain for \(\mathcal{H}_i\) and run the Markov chain until convergence. Then we predict the time and type for the event \(e_{i+1}\) conditional on \(\mathcal{H}_i\) the same as how we predict the event \(e_i\). 

\subsection{Time Complexity}\label{sec:training-time}
The time complexities for flip, swap, and resampling for a single event in MCMC are \(\mathcal{O}(1)\), constant irrespective of any values (amount of data, parameter values, length of time, etc).  A full analysis of the MCMC time complexity would require bounding the number of steps necessary (by mixing time arguments, usually). As with almost all other non-trivial MCMC inference methods, we do not have such a bound.

Intuitively, a model with more hidden layers has much larger search space for the events from a posterior distribution and it should take more time to get fully mixed. And we have observed experimental results to support our intuition. As shown in Table \ref{table:training-time}, the 2-hidden models take more time to get fully mixed than 1-hidden models.

The experiments are trained on a cluster with multiple CPU cores. The number of CPU cores ranges from 8 to 64.
\begin{table*}[htbp]
\caption{Training Time in Hours}
\label{table:training-time}
\vskip 0.15in
\begin{center}
\begin{small}
\begin{sc}
\begin{tabular}{lcr}
\toprule
Datasets & Model & \multicolumn{1}{r}{Time}\\
\midrule
\multirow{2}{12em}{Retweets} &1-hidden    & 12.0 \\
&2-hidden     & 108.5\\
\midrule
\multirow{2}{12em}{Earthquakes} &1-hidden    & 0.8         \\
&2-hidden     & 22.3 \\
\midrule
\multirow{2}{12em}{Homicides} &1-hidden    & 1.1         \\
&2-hidden     & 14.0 \\
\bottomrule
\end{tabular}
\end{sc}
\end{small}
\end{center}
\vskip -0.1in
\end{table*}

\section{LIMITATIONS} \label{sec:limitations}
\begin{itemize}
    \item The current gamma kernel function is simple, restricting the ability to model more complex data. In the future, we can try to design more complex kernels.
    \item We only have experiments for temporal point processes and no edge effects are taken into account here. However, a SPP is not constrained to one dimensional space. SPPs can exist in Euclidean spaces with any finite dimensions and  non-Euclidean spaces, like spheres and torus. Our next step would be to apply our method to more general SPPs.
\end{itemize}
\section{NEGATIVE SOCIETAL IMPACTS} \label{sec:neg-soc-impact}
Our model may be used to predict people's behavior from the data tracked by
smart phones or other wearable devices. Such predictions regarding when
people are away from their residence, visiting a medical facility, or at
work could be used for theft, medical rate increases, or overbearing
workplace monitoring.  Anonymization and privacy-preserving algorithms
could be used to mitigate such risks.

\end{document}